\documentclass[10pt,twocolumn,letterpaper]{article}

\usepackage[pagenumbers]{latex/cvpr} 

\usepackage{graphicx}
\usepackage{amsmath}
\usepackage{amssymb}
\usepackage{booktabs}
\usepackage{xcolor}         
\usepackage{multirow}
\usepackage{soul}
\usepackage{wrapfig}
\renewcommand\paragraph[1]{\vspace{0.12cm}\noindent\textbf{#1}}

\usepackage[pagebackref,breaklinks,colorlinks]{hyperref}

\usepackage[capitalize]{cleveref}
\crefname{section}{Sec.}{Secs.}
\Crefname{section}{Section}{Sections}
\Crefname{table}{Table}{Tables}
\crefname{table}{Tab.}{Tabs.}

\colorlet{dgreen}{green!50!black}
\colorlet{dred}{red!50!black}
\colorlet{dblue}{blue!70!black}

\newcommand{\modelname}{OVAD-Baseline }
\newcommand{\modelnamenospace}{OVAD-Baseline}
\newcommand{\datasetname}{OVAD}
\newcommand{\taskname}{OVAD}
\newcommand{\egupper}{\emph{E.g.}}
\newcommand{\nobjs}{14,300}
\newcommand{\natts}{1,401,484}
\usepackage{enumitem}
\setlist[enumerate]{leftmargin=2mm, label=\alph*)}
\setlist[itemize]{leftmargin=2mm}

\newcommand{\std}[1]{\textcolor{darkgray}{\tiny{$\pm$#1}}}

\def\cvprPaperID{7802} 
\def\confName{CVPR}
\def\confYear{2023}

\begin{document}

\title{Open-vocabulary Attribute Detection}

\author{Mar\'ia A. Bravo \qquad Sudhanshu Mittal \qquad Simon Ging \qquad Thomas Brox  \\ {\tt\small \{bravoma,mittal,gings,brox\}@cs.uni-freiburg.de} \\  University of Freiburg, Germany \\
\url{https://ovad-benchmark.github.io}}
\maketitle

\begin{abstract}
Vision-language modeling has enabled open-vocabulary tasks where predictions can be queried using any text prompt in a zero-shot manner. 
Existing open-vocabulary tasks focus on object classes, whereas research on object attributes is limited due to the lack of a reliable attribute-focused evaluation benchmark.
This paper introduces the Open-Vocabulary Attribute Detection~(OVAD) task and the corresponding OVAD benchmark.
The objective of the novel task and benchmark is to probe object-level attribute information learned by vision-language models. 
To this end, we created a clean and densely annotated test set covering 117 attribute classes on the 80 object classes of MS COCO. It includes positive and negative annotations, which enables open-vocabulary evaluation. Overall, the benchmark consists of 1.4 million annotations.
For reference, we provide a first baseline method for open-vocabulary attribute detection. Moreover, we demonstrate the benchmark's value by studying the attribute detection performance of several foundation models.
\end{abstract}

\section{Introduction}\label{sec:intro}
One of the main goals of computer vision is to develop models capable of localizing and recognizing an open set of visual concepts in an image. This has been the main direction for the recently proposed Open-Vocabulary Detection (OVD) task~\cite{ovr_baseline} for object detection, where the goal is to detect a flexible set of object classes that are only defined at test time via a text query.
Classical supervised object detection methods are bound to predict objects from a fixed set of pre-defined classes, and extending them to a very large number of classes is limited by the annotation effort.
OVD methods overcome this constraint by utilizing vision-language modeling to learn about novel objects using the weak supervision of image-text pairs.

OVD methods for object detection have made fast progress and have even surpassed supervised baselines for rare (tail) classes \cite{vild}. Best OVD methods \cite{vild, detic, regionclip, bridging} train with extra weak supervision using image classification datasets, which are focused on retrieving object information.
However, it is unclear on how well OVD methods generalize information beyond the object class. This paper focuses on object-level attribute information, such as the object's state, size, and color.
 


\begin{figure}[t]
\includegraphics[width=\linewidth]{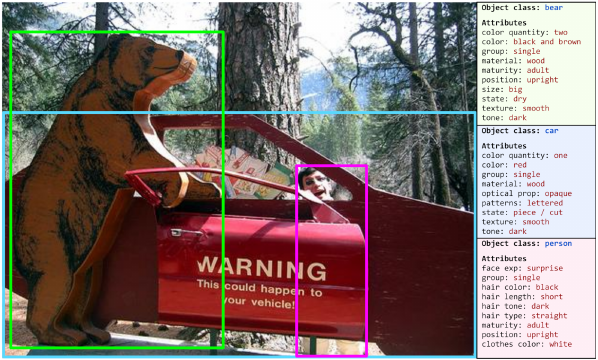}
\centering
\caption{Example from the presented open vocabulary attribute detection benchmark. The objective is to detect all objects and visual attributes of each object in the image. Objects and attributes are only specified at test time via text prompts.}
\label{fig:teaser_ovad}
\end{figure}

Attributes play a significant role in an object's identity. A small change of an attribute in a description can modify our understanding of an object's appearance and perception. 
Imagine driving in a forest where you encounter a bear like the one in Figure \ref{fig:teaser_ovad}. Even if you do not distinguish or know the type of bear, recognizing that it is made of wood is enough to realize that it is fake and harmless. 
A model capable of detecting object attributes enables a richer reasoning ability via combining objects and attributes. It allows the model to potentially extrapolate to novel object classes.

In this paper, we introduce the Open-Vocabulary Attribute Detection (\taskname) task.
Its objective is to detect and recognize an open set of objects in an image together with an open set of attributes for every object. 
Both sets are defined by text queries during inference without knowledge of the tested classes during training.
The \taskname\ task is a two-stage task.
The first stage, referred to as open-vocabulary object detection~\cite{ovr_baseline}, seeks to detect all objects in the image,
including \textit{novel} objects for which no bounding box or class annotation is available during training. 
The second stage seeks to determine all attributes present for each detected object. None of the attributes is annotated; therefore, all attributes are \textit{novel}.


Testing the \taskname\ task requires an evaluation benchmark with unambiguous and dense attribute annotations to identify misses as well as false positive predictions.
Current datasets~\cite{coco_attributes, vaw} for predicting attributes in-the-wild 
come with many missing or erroneous annotations, as discussed in more detail in Section~\ref{sec:ova_benchmark}.
Thus, in this paper, we introduce the \datasetname\ benchmark, an evaluation benchmark for open-vocabulary attribute detection. It is based on images of the MS COCO~\cite{coco} dataset and only contains visually identifiable attributes.
On average, the proposed benchmark has 98 attribute annotations per object instance, with 7.2 objects per image, for a total of 1.4 million attribute annotations, making it the most densely annotated object-level attribute dataset. 
It has a large coverage with 80 object categories and 117 attribute categories. 
It also provides negative attribute annotations, which enables quantifying false positive predictions.
The benchmark is devoid of various labeling errors since it is manually annotated and quality-tested for annotation consistency. 
Our \datasetname\ benchmark also extends the OVD benchmark~\cite{ovr_baseline} by including all 80 COCO object classes. This extension increases the novel set of objects from 17 to 32 classes. 
Together with the benchmark, we provide a first baseline method that learns the \taskname\ task to a reasonable degree. It learns the task from image-caption pairs by using all components of the caption, not only nouns. 
We also compare the performance of several off-the-shelf OVD models to get an insight of how much attribute information is implicitly comprised in nouns (e.g., \emph{puppy} implies a young dog).

Moreover, we demonstrate the value of the benchmark by evaluating object-level attribute information learned by several open-source vision-language models, sometimes also referred to as foundation models, including CLIP~\cite{clip}, Open CLIP~\cite{open_clip}, BLIP~\cite{blip}, ALBEF~\cite{albef}, and X-VLM~\cite{xvlm}. Such models learn from the weak supervision of image-text pairs, which is assumed to be available particularly via web content. The results show the extent to which the present success of foundation models on object classes generalizes to attributes.

\textbf{Contributions}
(1) We introduce the \textbf{O}pen-\textbf{V}ocabulary \textbf{A}ttribute \textbf{D}etection~(\taskname) task, 
where the objective is to detect all objects and predict their associated attributes. These objects and attributes belong to an open set of classes and can be queried using textual input.
(2) We propose the \datasetname\ benchmark: a clean and densely annotated evaluation dataset for open-vocabulary attribute detection, which can be used to evaluate open-vocabulary methods as well as foundation models.
(3) We provide an attribute-focused baseline method for the \taskname\ task, which outperforms the existing open-vocabulary models that only aim for the object classes.
(4) We test the performance of several open-source foundation models on visual attribute detection.

\section{Related Work}\label{sec:related_work}
\begin{figure*}[t!]
\centering
\begin{tabular}{cccc}
(a) Incorrect~\includegraphics[height=0.8em]{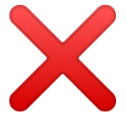} & 
(b) Missing~\includegraphics[height=0.8em]{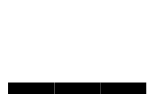} & 
(c) Ambiguous~\includegraphics[height=0.8em]{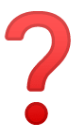} & 
(d) Non-visual~\includegraphics[height=0.9em]{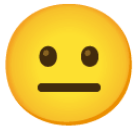}   \\
 \includegraphics[trim={0 0cm 0 0cm}, clip, width=35mm]{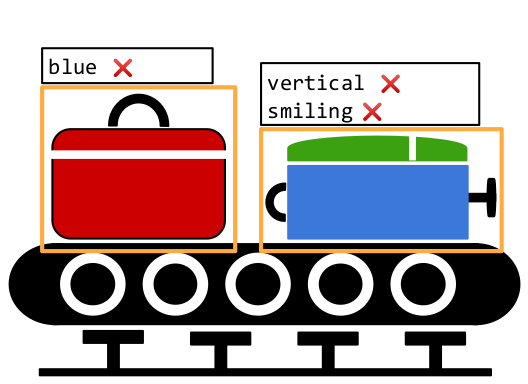} &
 \includegraphics[trim={0 0cm 0 0cm}, clip, width=35mm]{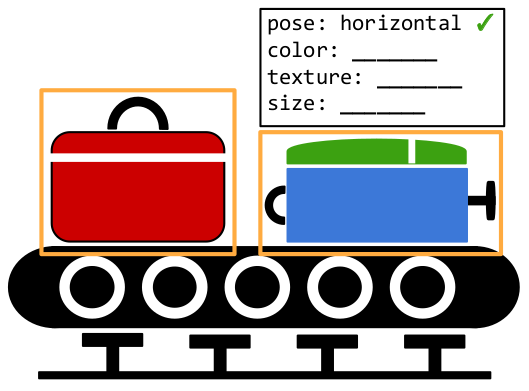} &
 \includegraphics[trim={0 0cm 0 0cm}, clip, width=35mm]{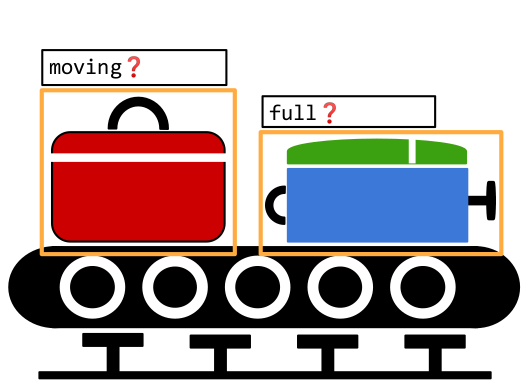} &
 \includegraphics[trim={0 0cm 0 0cm}, clip, width=35mm]{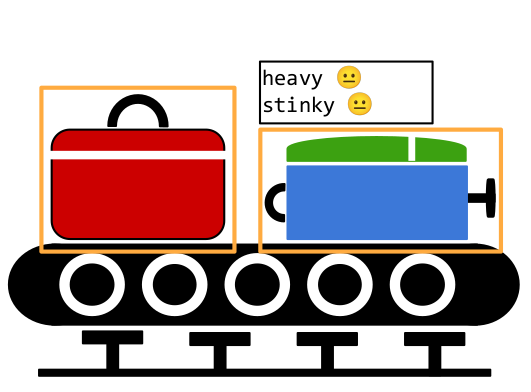} \\

 \includegraphics[trim={0 0.05cm 0 0cm}, clip, width=35mm]{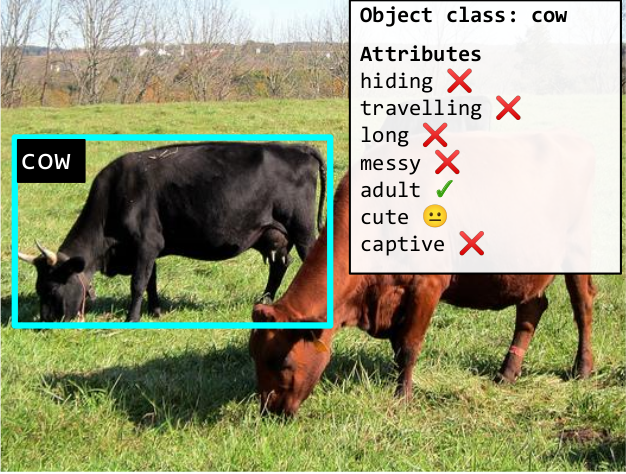} &
 \includegraphics[trim={0 0.05cm 0 0}, clip, width=35mm]{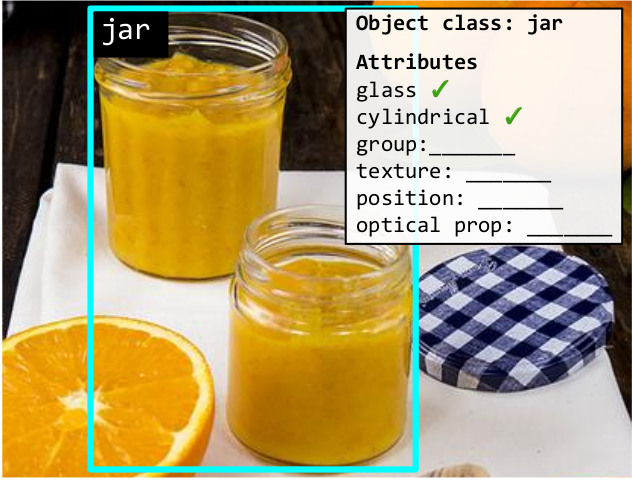} &
 \includegraphics[trim={0 0.1cm 0 0cm}, clip, width=35mm]{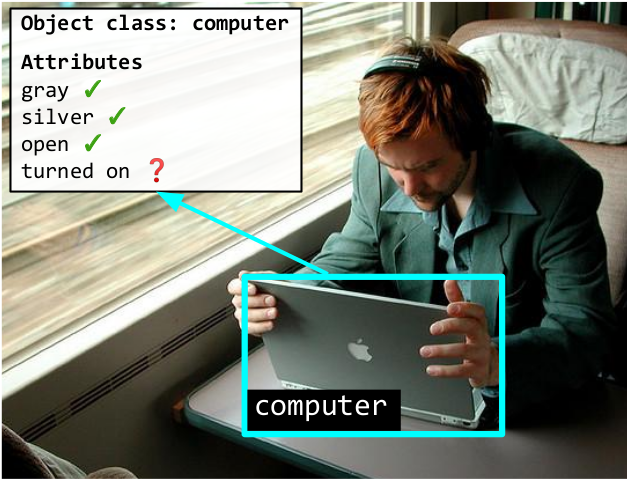} &
 \includegraphics[trim={0 0.0cm 0 0cm}, clip, width=35mm]{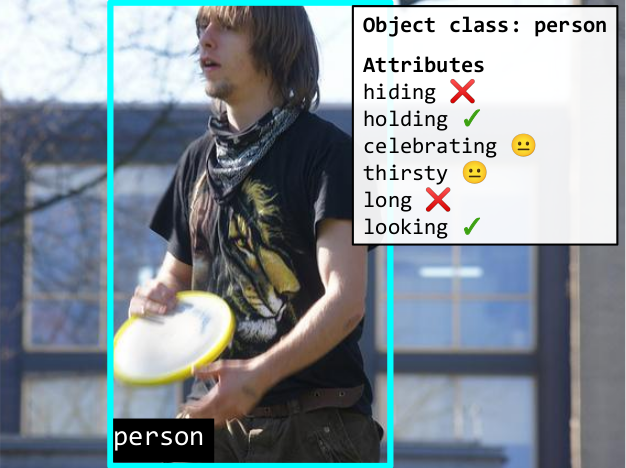}\\

 \includegraphics[trim={0 0.05cm 0 0cm}, clip, width=35mm]{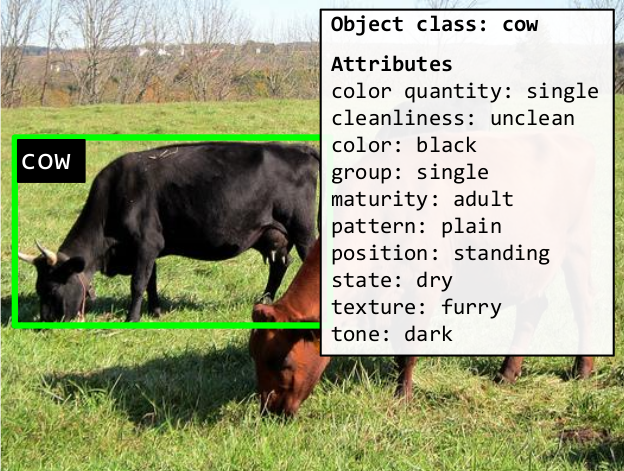} &
 \includegraphics[trim={0 0.05cm 0 0cm}, clip, width=35mm]{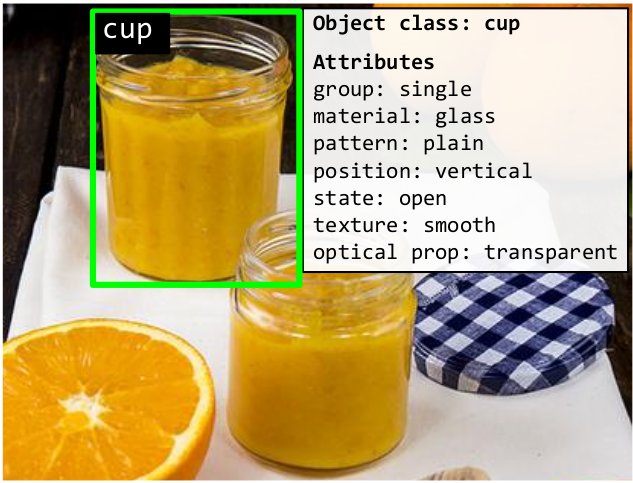} &
 \includegraphics[trim={0 0.1cm 0 0cm}, clip, width=35mm]{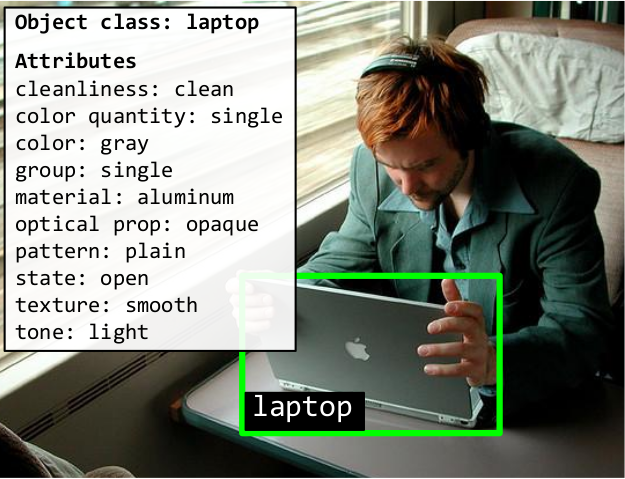} &
 \includegraphics[trim={0 0.0cm 0 0cm}, clip, width=35mm]{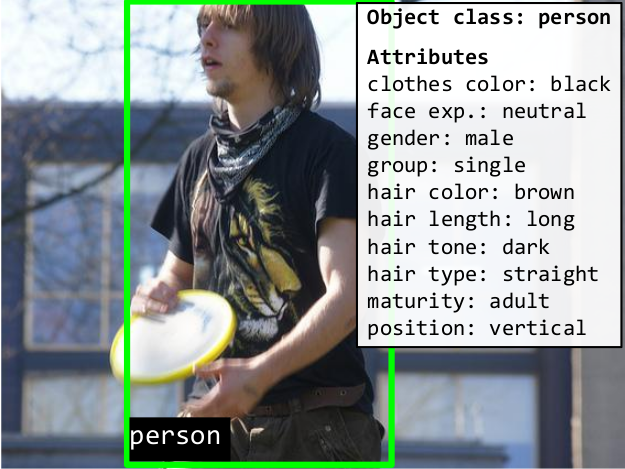} \\

\end{tabular}
\caption{Four major types of errors prominent in previous attribute benchmarks with examples and their improved version in the proposed benchmark (last row).
The top row shows a symbolic image with an example of how a briefcase and a trolley bag kept on a conveyor belt can be incorrectly marked with different types of errors. The second row of images shows examples from previous attribute benchmarks containing these errors. The last row shows examples from our proposed \datasetname\ benchmark.
}
\label{fig:error_types}
\end{figure*}

\paragraph{Attribute prediction} Several works have pursued the attribute prediction task to learn fine-grained information at different levels. 
Initial works focused on describing parts of the objects as attributes \cite{5206772, NIPS2007_ed265bc9}. 
In contrast to this partonomy identification, which can be regarded as a form of object detection (part detection), we focus on visual attributes represented by adjectives in human language. 
Other benchmarks for learning fine-grained semantics \cite{StatesAndTransformations, 6909426} 
focus on tasks within narrow class domains, such as
shoes~\cite{6909426}, clothes~\cite{han2017automatic, berg2010automatic}, birds~\cite{399}, and animals~\cite{xianCVPR17}. 
Another line of research \cite{NEURIPS2020_fa2431bf, sylvain2020locality, Al-Halah_2016_CVPR} focuses on zero-shot object classification by inferring the attributes of an object as an intermediate step or relying on object-attribute compositionality~\cite{Al-Halah_2016_CVPR, chen2020learning, li2020symmetry} for the task of zero-shot attribute-object classification. 
This work aims to evaluate the ability of vision-language models to detect and discriminate object-level attributes in a zero-shot manner.

\paragraph{Attribute detection benchmarks} Recent works predict
attributes in an open-domain setting, also known as ``in-the-wild'' setting. A few benchmarks have been proposed in this direction, along with some baseline methods. 
COCO Attributes \cite{coco_attributes} was the first such large-scale benchmark that annotated visual attributes for the COCO dataset. However, this dataset is limited in scope, with annotations only across 29 object categories.
Visual Genome~\cite{visual_genome} offers a much wider coverage of attribute categories with more than 68\,k attribute categories, including synonyms, but it contains very few attribute annotations for each object (0.74 attributes per instance). In Visual Genome, attribute annotations are not dense or exhaustive for every object since they were extracted from scene graph annotations which contain free-written form descriptions.
Its sparsity, noise, and lack of negative annotations make it unsuitable for evaluating the \taskname\ task. 
Other works have introduced visual question answering datasets~\cite{visual_genome, vqa} with questions that require an understanding of vision, language, and common sense to respond. 
Even though the answers to these questions overlap with our objective (\eg by asking about colors or materials), the performance on attributes and nouns cannot be isolated and analyzed using these datasets. 
VAW~\cite{vaw} proposed a large-scale dataset covering a wider range of attribute and object categories. They provide reliable positive and negative attribute labels for object instances and ensure that a minimum of 50 instances exist for each object-attribute pair. 
However, automated filtering techniques are used to keep the annotation cost feasible, resulting in very sparse annotations in terms of the number of instances per image and attributes per instance. 
Open Images~\cite{open_images} is a dataset consisting of 9 million images with image-level labels and bounding boxes. It provides attribute annotations for 288 object categories; however, it is limited to only 15 attribute categories that are not densely annotated for each object.
We find that these benchmarks are of limited use for the precise evaluation and analysis of \taskname\ task. 
Therefore, in this work, we propose a new evaluation benchmark for attribute detection with clean and dense attribute annotations. 

\paragraph{Open-vocabulary methods} 
Zarenian~\etal~\cite{ovr_baseline} introduced the open-vocabulary object detection problem, where the goal is to detect an open set of classes, some annotated (base classes) during training and others only defined at test time (novel classes).
In this setting, the model learns in a weakly-supervised manner using image-caption pairs along with the annotations of base object classes. Various follow-up works \cite{locov, vild, detic, bridging, pb-ovd} have improved the performance of open-vocabulary object detection. Bravo~\etal~\cite{locov} proposed a localized image-caption matching technique. Gu~\etal~\cite{vild} proposed an improved model using a pre-trained open-vocabulary classification model~\cite{clip}, created a new benchmark on LVIS~\cite{lvis}, and showed some initial qualitative examples of fine-grained object detection by adding adjectives to the object queries. Recently, Zhou~\etal~\cite{detic} trained the classifier module of the detector by using extra class annotations. Other works ~\cite{bridging, pb-ovd, vl-plm} used pseudo-bounding-box annotations of base and novel classes to train their detector. In this work, we expand this problem formulation to include attributes.

\paragraph{Vision-language models} have changed the way of approaching semantic learning tasks in computer vision by enabling the usage of large-scale free annotated data from the web.
These foundation models \cite{clip, albef, open_clip, blip, vilt, meter, align, xvlm} use cross-modal objectives to learn to align visual concepts to their language representation leading them to achieve state-of-the-art performance on visual reasoning tasks.
In this paper, we challenge five state-of-the-art vision-language models on the fine-grained task of open-vocabulary attribute detection.

\section{Open-vocabulary Attribute Detection}\label{sec:dataset}
\subsection{The \taskname\  Task}\label{sec:ovad_task}

Open-vocabulary attribute detection has a two-fold objective: (1) object detection and (2) discovery of attributes for all detected objects. 
Both object detection and attribute detection are formulated as open-vocabulary tasks. The first is known as open-vocabulary detection (OVD).

In previous work~\cite{ovr_baseline}, OVD considers two disjoint sets of object classes - base $\mathcal{O}^B$ and novel $\mathcal{O}^N$ classes. 
The class labels and bounding boxes are given for the first set $\mathcal{O}^B$ during training, whereas the second set $\mathcal{O}^N$ needs to be derived automatically from image-caption pairs. Only at test time the set $\mathcal{O}^N$ is revealed. To be compatible with this setting from the literature, we use the object detection part of the \taskname\ task in the same way. 

In contrast, for the second objective of \taskname\ task, none of the attributes are known during training. Rather all knowledge about attributes must be derived from image-caption pairs or pretrained vision-language models. Only at test time, the set of tested-visual attributes $\mathcal{A}$ is revealed. Using knowledge about the tested set of attribute classes for building the model violates the definition of the task. 

Solving the task of \taskname\ requires the ability to detect both $\mathcal{O}^B$ and the unbounded $\mathcal{O}^N$ set of object classes as well as to determine whether attributes from $\mathcal{A}$ are present or absent for every object. 

We also provide the \taskname\ task in a box-oracle setting, where the bounding box and object class annotations are available for all objects during inference. Thus, we only evaluate the second objective of the multi-label attribute detection task. This setting evaluates the attribute detection in isolation, independent of the mistakes made in the object detection part. 

\subsection{The \datasetname\ Benchmark}
\label{sec:ova_benchmark}
For evaluating \taskname, it is necessary to have a benchmark dataset that contains annotations of both objects $\mathcal{O}$ and attributes $\mathcal{A}$. 
First, we discuss the limitations of previous datasets that provide both object and attribute annotations and then show how we rectify them for our benchmark.

\paragraph{Types of errors} We identify four major sources of annotation errors in previous datasets, which make them unsuitable for the \datasetname\ benchmark. The boundaries between these error types are blurry. 
Figure~\ref{fig:error_types} shows an example for each error type followed by our corrected version to give an intuition for each of them. We summarize them as follows:  

\begin{itemize}
\item \textbf{Type-A Incorrect:} Objects with incorrect attribute annotations. As shown in Figure~\ref{fig:error_types}(a), the \textit{cow} is marked incorrectly with \textit{hiding, travelling, long}, etc.

\item \textbf{Type-B Missing:} Objects missing attribute annotations. As shown in Figure~\ref{fig:error_types}(b), the \textit{jar} has missing attributes such as \textit{group, texture}, and \textit{position}.

\item \textbf{Type-C Ambiguous:} Attributes that cannot be marked using the given image due to incomplete information.

Figure~\ref{fig:error_types}(c) shows a \textit{bag} on the conveyor belt marked as \textit{moving}, and a \textit{computer} marked as \textit{turned on}, in the top and middle row respectively. These attributes only become valid when considering temporal information or a front view of the computer. 

\item \textbf{Type-D Non-visual:} Attributes that cannot be marked using visual information. These attributes are often subjective such as certain emotions or states of mind and occur due to poor selection of the attribute set.
As shown in Figure~\ref{fig:error_types}(d), the \textit{person} is annotated as \textit{celebrating} and \textit{thirsty}. 

\end{itemize}

We aim to overcome the above-mentioned limitations of previous datasets by selecting a good set of attribute classes that can be accurately annotated for all object categories and is visually non-ambiguous for most samples. 
Our \datasetname\ evaluation benchmark comprises 2000 images randomly selected from the MS-COCO \cite{coco} validation set. To ensure a densely annotated dataset with a large number of object annotations in an image, we started our annotation process with the COCO~\cite{coco} object detection benchmark. We added bounding boxes for missing objects, revised inaccurate boxes, and removed incorrect object annotations. 
As a result, we obtained \nobjs\ object instances for the attribute annotation process.
We manually labeled each object instance with 117 attributes following strict annotation guidelines to avoid above mentioned errors. 
The \datasetname\ benchmark dataset is designed as a test set to evaluate models' fine-grained open-vocabulary detection capabilities. It is neither designed for classical supervised training nor as a validation set, as both contradict the open-vocabulary paradigm. \\

\paragraph{Selection of attributes}
We extracted adjectives from the captions of the COCO Captions dataset~\cite{cocoCaptions} using a parts-of-speech detector~\cite{nltk}. We selected the adjectives that occurred at least ten times and grouped them by synonyms using WordNet~\cite{wordnet}, Collins English Dictionary, and Oxford English Dictionary. 
We retained the synonyms and manually removed abstract, action-based, and non-visual attributes, such as \textit{peaceful, walking, thirsty, etc.}, as shown in Figure~\ref{fig:error_types}(c\&d). 
We considered the 80 MS-COCO object classes and removed attribute classes for which no positive object-attribute example existed. After this process, our final set consists of 117 unique attribute categories. 
We built a taxonomy and identified 19 attribute types or superclasses corresponding to \textit{color, pattern, material, maturity, cooking state} and 14 others. A detailed diagram of the attribute taxonomy is included in the supplementary.

\paragraph{Annotation process} The \datasetname\ benchmark is fully annotated by humans, as compared to other works~\cite{coco_attributes, vaw, visual_genome}. This ensures accurate ground-truth labels. The annotation was done using the open-source annotation platform ``Computer Vision Annotation Tool" (CVAT)~\cite{cvat}. The \datasetname\ benchmark has all attributes marked either as \textit{positive}, \textit{negative}, or \textit{unknown}. We use the attribute taxonomy and the attribute types during the annotation process. Most of the attributes are mutually exclusive within their attribute type, \eg, \textit{pose} can be either \textit{vertical} or \textit{horizontal} but not both simultaneously. For every object, annotators were directed to select one of the attributes for every attribute type as positive or unknown. Given the exclusiveness property, all non-selected attributes within the same attribute type were marked as negatives or unknown, respectively. 
This produced dense annotations for every object and ensured that the missing type errors were diminished (see Figure~\ref{fig:error_types}(b)). 
Attributes marked as unknown are excluded during evaluation. The unknown option either refers to an unknown attribute for an instance or an in-between case, where a discrete label can not be assigned clearly. This helps rectify ambiguous type errors like in Figure~\ref{fig:error_types}(c). 
We manually excluded infeasible object-attribute combinations during annotation, such as \textit{smiling cup} or \textit{open person}, to avoid incorrect type errors shown in Figure~\ref{fig:error_types}(a) and speed up the annotation process. 
We include a detailed description of the annotation process in the supplementary.

\begin{table}[t]\scriptsize
\begin{center}
\begin{tabular}{|@{\hspace{1.5mm}}l@{\hspace{1.5mm}}|@{\hspace{1.5mm}}c@{\hspace{1.5mm}}|@{\hspace{1.5mm}}c@{\hspace{1.5mm}}|@{\hspace{1.5mm}}c@{\hspace{1.5mm}}|@{\hspace{1.5mm}}c@{\hspace{1.5mm}}|}
    \hline
    \textbf{Dataset}     & \textbf{\datasetname} (ours)      &  \textbf{VAW}~\cite{vaw}      & \textbf{COCO-A}~\cite{coco_attributes}       & \textbf{VG}~\cite{visual_genome}
    \\  
    Purpose     & Test & Train+Test & Train+Test & Train+Test
    \\  
    \hline  
    \multicolumn{5}{|l|}{ \textit{\# categories} }\\
    \hline  
    Objects     & 80                & 2,260     & 29                & 33,877    \\
    Attributes  & 117               & 620       & 196               & 68,111       \\
    Negative Labels  & Yes          & Yes       & No                & No \\
    \hline    
    \multicolumn{5}{|l|}{ \textit{\# instances} }\\
    \hline  
    Objects     & \nobjs            & 260,895   & 188,426           & 3.8M      \\
    Attribute   & 1.4M              & 0.9M      & 3.4M              & 2.8M      \\
    Images      & 2,000             & 72,274    & 84,044            & 108,077   \\
    \hline    
    \multicolumn{5}{|l|}{ \textit{\# instances per image} }\\
    \hline  
    Objects     & 7.2               & 3.6       & 2.2                   & 35  \\
    \multirow{2}{*}{Attributes}  & 700.7             & 12.83     & \multirow{2}{*}{41.08}                 & \multirow{2}{*}{26}  \\
                & (+)61   (-)639    & (+)5.4   (-)7.4 &  &  \\
    \hline    
    \multicolumn{5}{|l|}{ \textit{\# instances per box} }\\
    \hline  
    \multirow{2}{*}{Attributes}  & 96.8              & 3.56      & \multirow{2}{*}{18.33}                 & \multirow{2}{*}{0.74} \\
                & (+)8.3   (-)88.5  & (+)1.51   (-)2.05 &                       &  \\
    
    \hline  
\end{tabular}
\caption{Statistics of object-level attribute benchmarks. \datasetname\ is densely annotated as compared to other datasets. (+) and (-) indicate positive and negative attribute labels respectively.
}
\label{tab:datasets}
\end{center}
\end{table}

\begin{figure*}[t]
\includegraphics[width=0.8\linewidth]{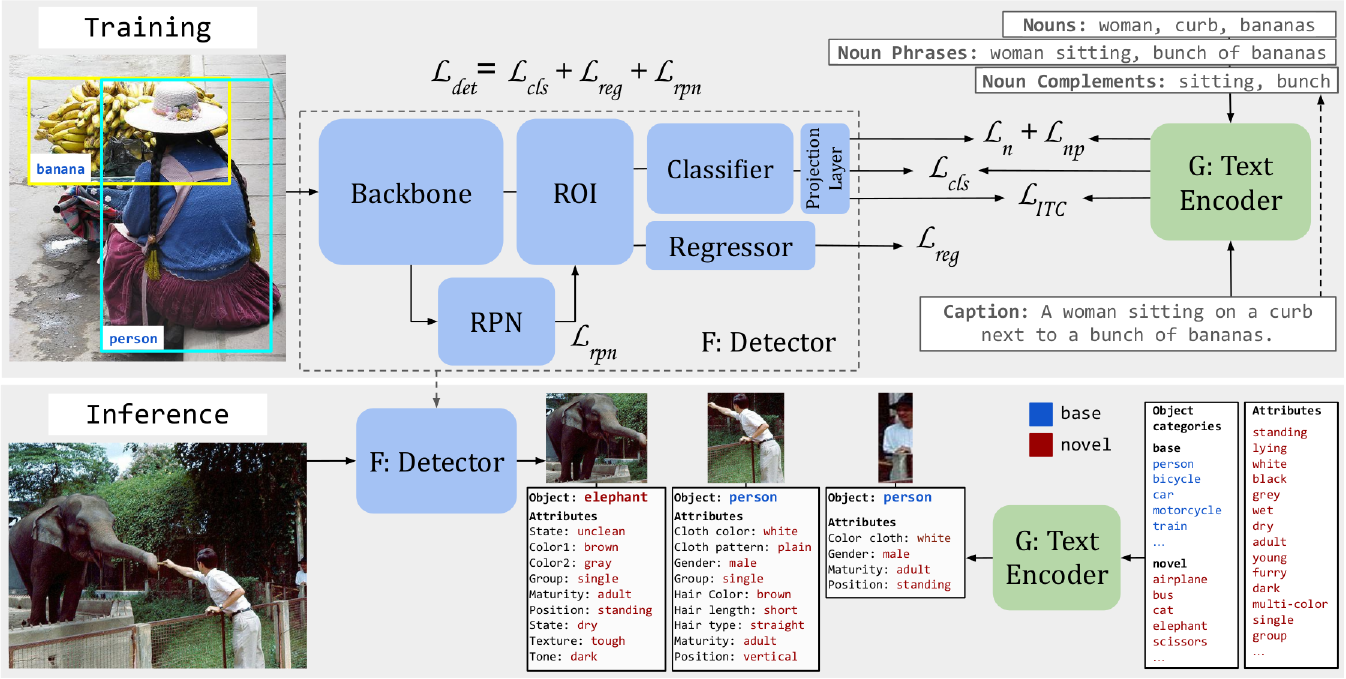}
\centering
\caption{Training and inference setup for the \modelname model. The method is a two-stage detector that matches image regions with text embeddings of nouns, noun phrases, noun complements, and captions. At inference, the detector detects the \textcolor{blue}{base} and \textcolor{dred}{novel} objects and their attributes by matching box-region embeddings with text embeddings of the object and attribute classes. 
}
\label{fig:model}
\end{figure*}
 
\paragraph{Statistics} 
The \datasetname\ benchmark is a medium-scale benchmark with a total of \natts\ attribute annotations over 2000 images. It considers 117 attribute categories that span across 80 object categories with a total of \nobjs\ object instances.
There are 122,998 positive and 1,278,486 negative attribute annotations in total and 172,760 attribute instances are marked as unknown. 
Table~\ref{tab:datasets} shows a summary of the dataset statistics together with other attribute datasets. Since \datasetname\ is exclusively an evaluation benchmark, the number of images is not comparable to the other datasets.
The \datasetname\ evaluation benchmark is densely annotated with 7.2 box annotations per image, compared to 3.6 instances per image in VAW. Our benchmark offers, on average, 96.8 attribute annotations per box, with a total of 700.7 attribute annotations per image. 
This is much larger than any other object-level attribute benchmark. The benchmark provides both positive and negative attribute annotations grouped into 19 types of attributes.

\paragraph{Evaluation metric} 
As discussed in Section~\ref{sec:ovad_task}, the \taskname\ task can be evaluated under two settings: (1) open-vocabulary detection and (2) box-oracle setting. 
In the open-vocabulary detection setting, each ground-truth object instance is matched with at most one object prediction. To qualify as a positive match, the detection must have an Intersection over Union (IoU~\cite{pascal_voc}) $\ge 0.5$ independent of the ground-truth class. 
For every ground-truth object, the prediction with maximum IoU overlap is considered as the matching predicted object. We evaluate attribute performance by comparing the attribute scores and labels of matching ground-truth and predicted objects. 
Following Veit~\etal~\cite{veit2017learning}, in the case that a ground-truth object has no matching prediction (IoU $< 0.5$ for all predictions), all attributes are marked as absent.
We calculate the average precision (AP)~\cite{pascal_voc} for every attribute category independently and then average across categories (mAP)~\cite{pascal_voc}.
Additionally, for completeness, we evaluate mAP at $0.5$ IoU for open-vocabulary object detection on the 80 class object set; we call this set OVD-80. We use the \textit{Generalized} evaluation that considers the probability across all object classes (base and novel). 
In the box-oracle setting, the attribute mAP metric is directly evaluated for ground-truth bounding boxes in an object-class-agnostic manner. 

\section{OVAD Baseline Method}\label{sec:method}
In this section, we provide a baseline method for the \taskname\ task.
The objective is to learn a vision model that detects objects and their corresponding attributes in an open-vocabulary manner.  
Our \modelname comprises two models: a frozen language model $G$ and an object detector $F$ based on Faster-RCNN~\cite{faster_rcnn}, where we replace the classification head with a linear layer that projects the visual features to the language space produced by $G$. Following other works~\cite{detic, vild, regionclip, bridging}, we use CLIP~\cite{clip} as the language model. We define $g_w = G(w)$ as the embedding representation of a text composed of one or more words $w$, and $f_b=F(I_b)$ as the embedding representation of a box-region $b$ of an image $I$.\\

\paragraph{Visual-text matching} Throughout the paper, we use image-text pairs for learning the vision language alignment. These pairs can correspond to images and captions, box-regions and class labels, or in a more general setting, any box-region and text. We use the cosine similarity
\begin{equation}\label{eq:similarity}
    s_{w,b}=\sigma(\frac{g_w \cdot f_b}{|g_w||f_b|} \cdot \tau)
\end{equation}
as matching score between a text $w$ and a box-region $b$, where $\tau$ is a temperature hyper-parameter and $\sigma$ corresponds to the sigmoid function. 

\paragraph{Training objectives} The detector $F$ is trained with three objectives: 1) learn to localize objects in an image, 2) semantically match image representations with caption embeddings, and 3) train the classifier branch with proxy-labels to predict the novel classes and attributes.

For the first objective, we train $F$ with labels and bounding box coordinates of the base classes $O^B$. 
We use the standard detection loss $\mathcal{L}_{det}$ from Faster R-CNN (shown in Figure~\ref{fig:model}) adapted for open-vocabulary. 
It comprises three losses: a region proposal network loss $\mathcal{L}_{rpn}$~\cite{faster_rcnn}, a class-agnostic $l_1$ loss as box regression loss $\mathcal{L}_{reg}$, and a similarity-based classification loss $\mathcal{L}_{cls}$ using the binary cross-entropy loss over the similarly score \eqref{eq:similarity} between the visual embedding of the object box and the text embedding of the base classes.

For the second objective we use the image-text contrastive matching (ITC) loss
\begin{equation}\label{eq:bce_loss}
    \mathcal{L}_{ITC} = -{(y\log(s_{C,I}) + (1 - y)\log(1 - s_{C,I}))},
\end{equation}
with $s_{C,I}$ being the similarity score \eqref{eq:similarity} between the image $I$ and the caption $C$, and $y \in \{1, 0\}$ depending on whether $I$ and $C$ are a positive pair.
We apply this loss to positive and negative image-caption pairs. 

For the third objective, we match concepts within captions with image regions.
These concepts, referred to as `parts-of-caption' in this work, include nouns, noun phrases, and noun complements. They act as proxy-labels for objects and attributes. 
We obtain these parts-of-caption using a part-of-speech tagging method from the open-source software spaCy~\cite{spacy}.
Nouns usually refer to object classes; however, they often reveal some attribute information, 
\eg, \textit{man/woman} are nouns that reveal gender, \textit{cows} is a plural noun that reveals the quantity attribute. Noun phrases are usually adjective-noun combinations, which contain more explicit attribute information, such as~\textit{red helmet, wooden table}. 
We remove the nouns from the noun phrases to obtain ``noun complements'', which often contain adjectives, and use these to match directly with image regions. 
Since the location of these parts-of-caption is unknown, we match proxy-labels with the biggest predicted bounding box features $F(I_{b_{max}})$, similar to the usage of image labels in Detic's~\cite{detic} training. 
Along with these positive pairs, we create negative proxy-labels using arbitrary image-caption pairs and apply the binary cross entropy loss~\eqref{eq:bce_loss}. We refer to these losses as $\mathcal{L}_{n}$ and $\mathcal{L}_{np}$ for nouns and noun phrases/complements, respectively. 

\paragraph{Inference} During inference time, we consider a vocabulary composed of all object classes, $\mathcal{O}^B \cup \mathcal{O}^N$, together with the attribute classes $\mathcal{A}$ and use the language model $G$ to get the corresponding text-vector representations of every class, as shown in Figure~\ref{fig:model}. 
We do not use any special text prompt for this purpose but consider all synonyms for every class (object/attribute) and average their text-vector representations. 
We obtain the final prediction for object and attribute classes by taking the sigmoid of the similarity~\eqref{eq:similarity} between the box-region representation $F(I_b)$ and the class-text embedding $G(c)$. We compute the output separately for each object and attribute class, predicting the class' presence or absence. 
See the supplementary for implementation details.


\begin{table}[t]\scriptsize
\centering 
\begin{tabular}{ |@{\hspace{0.5mm}}c@{\hspace{0.2mm}}|@{\hspace{0.5mm}}c@{\hspace{0.5mm}}|@{\hspace{0.5mm}}c@{\hspace{0.5mm}}c@{\hspace{0.5mm}}c@{\hspace{0.5mm}}|@{\hspace{0.2mm}}c@{\hspace{0.5mm}}c@{\hspace{0.5mm}}c@{\hspace{0.5mm}}|} 
    \hline 
    \multirow{2}{*}{\textbf{Method}} & \multicolumn{4}{c}{ \textbf{\taskname} } & \multicolumn{3}{c|}{ \textbf{Generalized OVD-80} }  \\
    & All & Head & Medium & Tail
    & Novel \tiny{(32)} & Base \tiny{(48)} & All \tiny{(80)}\\ 
    \hline
    Chance       & 8.6 & 36.0 & 7.3 & 0.6 & - & - & - \\ %
    \hline
    OV-Faster-RCNN    & 11.7 & 34.4 & 13.1 & 1.9 & 0.3 & 53.3 & 32.1\\ %
    VL-PLM~\cite{vl-plm}  & 13.2 & 32.6 & 16.3 & 2.6 & 19.7 & 58.8 & 43.2 \\
    Detic~\cite{detic}        & 13.3 & 44.4 & 13.4 & 2.3 & 20.0 & 49.2 & 37.5\\
    Rasheed \etal~\cite{bridging}     & 14.6 & 33.5 & 18.7 & 2.8 & 32.5 & 56.6 & 46.9\\
    LocOv~\cite{locov}        & 14.9 & 42.8 & 17.2 & 2.2 & 22.5 & 52.5 & 40.5\\ %
    OVR~\cite{ovr_baseline}   & 15.1 & 46.3 & 16.7 & 2.1 & 17.9 & 51.8 & 38.2\\
    \modelname & \textbf{18.8\std{0.3}} & \textbf{47.7\std{0.6}} & \textbf{22.0\std{0.5}} & \textbf{4.6\std{0.5}} & 24.7\std{0.6} & 49.1\std{0.2} & 39.3\std{0.4}\\ %
    \hline
\end{tabular}
\caption{mAP for Open-vocabulary Attribute Detection (OVAD) and AP\textsubscript{50} on Open-Vocabulary Detection (OVD-80).}
\label{tab:ovad_sota_gen}
\end{table}

\section{Experiments}\label{sec:exp}

\begin{table}[t]\scriptsize
    \centering
\begin{tabular}{ |c@{\hspace{2.2mm}}c@{\hspace{2.2mm}}c@{\hspace{2.2mm}}c@{\hspace{2.2mm}}c|c|c| } 
 \hline
 box+cls & \multirow{2}{*}{captions} & \multirow{2}{*}{nouns} & noun & noun & OVAD & AP\textsubscript{50} - OVD-80\\
 $\mathcal{O}^B$ & & & phrases  & comp. & mAP  & Novel (32) \\
 \hline
 \checkmark & & & & & 11.7\std{0.1} & 0.3\std{0.3} \\ 
 \checkmark & \checkmark & & & & 15.0\std{0.2} & 19.2\std{0.1} \\ 
 \checkmark & \checkmark & \checkmark & & & 16.2\std{0.3} & 23.2\std{0.8} \\ 
 \checkmark & \checkmark & \checkmark & \checkmark & & 15.9\std{0.1} & 23.7\std{0.5} \\ 
 \checkmark & \checkmark & \checkmark & & \checkmark & 18.8\std{0.3} & 24.7\std{0.6} \\ 
 \hline
\end{tabular}
\caption{Text input ablation. OVAD and OVD-80 performance on novel classes using different types of text granularity as proxy-labels to train the model. \textit{box+cls}: box and object-class labels for base objects, \textit{noun phrases}: phrases that have one noun and some modifiers, \textit{noun compl.}: noun phrases without the main noun. Training with finer granularity of text supervision is favorable.
}
\label{tab:text_ablation}
\end{table}

\begin{table*}[t]
\begin{minipage}[lt]{0.6\textwidth}{\scriptsize
\centering 
\begin{tabular}{ |c|@{\hspace{0.3mm}}c@{\hspace{0.3mm}}|c|c@{\hspace{0.5mm}}c@{\hspace{0.5mm}}c|} 
    \hline 
    \multirow{2}{*}{\textbf{Method}} & \textbf{Training} & \multicolumn{4}{|c|}{ \textbf{\taskname-Box} } \\
    & \textbf{Data} & All & Head & Medium & Tail \\ 
    \hline
    Chance & - & 8.6 & 36.0 & 7.3 & 0.6 \\
    \hline
    CLIP RN50~\cite{clip}  & 400M (9) & 15.8 & 42.5 & 17.5 & 4.2 \\
    CLIP VIT-B16~\cite{clip}  & 400M (9) & 16.6 & 43.9 & 18.6 & 4.4 \\
    \hline
    Open CLIP RN50~\cite{open_clip}  & 12M (7b) & 11.8 & 41.0 & 11.7 & 1.4 \\
    Open CLIP ViT-B16~\cite{open_clip}  & 400M (8b) & 16.0 & 45.4 & 17.4 & 3.8 \\
    Open CLIP ViT-B32~\cite{open_clip}  & 2B (8c) & 17.0 & 44.3 & 18.4 & 5.5 \\
    \hline
    ALBEF~\cite{albef}  & 4M (1a,3,4,7a) & 15.6 & 43.1 & 17.3 & 3.7 \\
    ALBEF~\cite{albef}  & 14M (1a,3,4,7) & 15.3 & 43.7 & 17.1 & 3.0 \\
    ALBEF~\cite{albef}  & 14M (1a,3,4,7) + ft(2) & 21.0 & 44.2 & 23.9 & 9.4 \\
    \hline
    BLIP~\cite{blip}  & 14M (1a,3,4,7) & 17.0 & 46.6 & 18.3 & 5.0 \\
    BLIP~\cite{blip}  & 129M (1a,3,4,7,8a) & 18.2 & 44.4 & 20.7 & 5.7 \\
    BLIP~\cite{blip}  & 129M (1a,3,4,7,8a) + ft(1a) & 24.3 & 51.0 & 28.5 & 9.7 \\
    \hline
    X-VLM~\cite{xvlm}  & 4M (1$^{*}$,3$^{*}$,4,7a) & 25.9 & \textbf{50.3} & 32.0 & 9.8 \\
    X-VLM~\cite{xvlm}  & 16M (1$^{*}$,3$^{*}$,4,5$^{*}$,6$^{*}$,7) + ft(2) & 26.2 & 48.7 & 31.2 & 12.1 \\
    X-VLM~\cite{xvlm}  & 16M (1$^{*}$,4$^{*}$,4,5$^{*}$,6$^{*}$,7) & \textbf{28.1} & 49.7 & \textbf{34.2} & \textbf{12.9} \\
    \hline
    \modelnamenospace{}-Box  & 0.11M (1a,1b$^{*\text{base}}$) & 21.4\std{0.4} & 48.0\std{0.5} & 26.9\std{0.6} & 5.2\std{0.5} \\
    
    \hline
\end{tabular}
\caption{Open-vocabulary Attribute Detection results (mAP) for foundation models in the box-oracle setup (\taskname-Box).  $^{*}$ The model uses the localization information in the annotations of this dataset. + ft: final fine-tuning pass on the captions of this dataset. Table~\ref{tab:training_datasets} details the training datasets.}
\label{tab:ova_sota_box}}
\end{minipage}
\begin{minipage}[rt]{0.40\textwidth}{\scriptsize
\centering 
\begin{tabular}{|@{\hspace{0.9mm}}r@{\hspace{0.6mm}}l@{\hspace{0.9mm}}|@{\hspace{0.9mm}}c@{\hspace{0.9mm}}c@{\hspace{0.9mm}}c@{\hspace{0.9mm}}c@{\hspace{0.9mm}}|}
    \hline 
    \textbf{(\#)} & \textbf{Dataset}         & \textbf{\#Images}  & \textbf{\#Captions}    & \textbf{\#Objects}     & \textbf{\#Regions} \\
    \hline 
    (1a) & COCO Captions~\cite{cocoCaptions}    & 0.12M   & 0.57M       & -             & -         \\
    (1b) & COCO Objects~\cite{coco} & 0.12M   & -           & 0.86M         & -         \\
    (2) & RefCOCO+~\cite{refcoco}        & 0.019M  & -           & -             & 0.14M   \\
    (3) & VG~\cite{visual_genome}   & 0.10M   & -           & 2.5M          & 5.4M      \\
    (4) & SBU Captions~\cite{sbu_captions}   & 1M      & 1M          & -             & -         \\
    (5) & OpenImages~\cite{open_images}      & 1.7M    & 0.67M       & 4.4M          & 3.3M      \\
    (6) & Objects365~\cite{objects365}      & 1.8M    & -           & 29M           & -         \\
    (7a) & CC-3M~\cite{cc3m}          & 2.95M   & 2.95M       & -             & -         \\
    (7b) & CC-12M~\cite{cc12m}         & 11.1M   & 11.1M       & -             & -         \\
    (8a) & LAION~\cite{laion400m}         & 115M    & 115M        & -             & -         \\
    (8b) & LAION~\cite{laion400m}         & 400M    & 400M        & -             & -         \\
    (8c) & LAION~\cite{laion2b}         & 2B      & 2B          & -             & -         \\
    (9) & CLIP 400M~\cite{clip}      & 400M    & 400M        & -             & -         \\
    \hline 
\end{tabular}
\caption{Training set legend and statistics}
\label{tab:training_datasets}}
\end{minipage}
\end{table*}
\subsection{Open-vocabulary Attribute Detection} 
\paragraph{Open-vocabulary baseline methods}
We compare our \modelname with previous off-the-shelf OVD models. For all methods base class object annotations come from MS COCO~\cite{coco} 2017 training set and caption annotations from COCO Captions~\cite{cocoCaptions} 2017 training set. 
Given that OVD methods project the visual information to a language space, we use the similarity \eqref{eq:similarity} of the visual representation of detected objects and the text embedding of every attribute to produce the attribute predictions, similar to the inference in Figure~\ref{fig:model}. 

OV-Faster-RCNN is a Faster-RCNN adapted for open-vocabulary. Similar to \modelnamenospace, the classification head of the detector is replaced with a linear layer to project the visual representation to the language space from the CLIP~\cite{clip} text encoder. We train the detector network only using the class names of the base object classes and their box annotations. No caption was used for training. 

OVR~\cite{ovr_baseline} and LocOv~\cite{locov} train the object detector using two stages. First, the detectors learn a mapping between image regions and tokens in the caption via attention-based image-caption matching. OVR uses image grid-based regions for the matching, whereas LocOv introduces additional object proposal boxes.
In the second stage, the models are fine-tuned using the base class annotations to learn the object detection task. Both models use BERT~\cite{bert} as the text encoder.

Detic~\cite{detic} and Rasheed \etal~\cite{bridging} 
train the detector using image-level labels filtered from the captions. 
Labels correspond to objects and are filtered using the class names of both base and novel classes, which technically is closed-vocabulary.
Detic matches image-level labels, in text format, with the biggest box proposal. Rasheed \etal~\cite{bridging} first produce pseudo-labels for box proposals, using the image-level labels, to train the classification head of the detector. Similarly, VL-PLM~\cite{vl-plm} uses CLIP scores and from a class-agnostic object proposals to get pseudo-labels and train the OVD. All three models use CLIP~\cite{clip} as the text encoder.

\paragraph{Results on the \datasetname\ benchmark} 
Table \ref{tab:ovad_sota_gen} presents results on the proposed \datasetname\ benchmark for the six open-vocabulary detection methods. It shows results for attribute detection (\taskname) and object detection (OVD-80). Given that the attribute frequency has a long-tailed distribution and following previous works ~\cite{lvis, vaw}, we report separate performances on attributes in the `head', `medium', and `tail' of this distribution. 
These sets contain 16, 55, and 46 classes, respectively (see the supplementary for details).

All methods yield results above the chance level, even though the OVD methods were not designed to recognize attributes but only objects. Our \modelname method outperforms these OVD methods. 
Methods that match image-regions with text-parts, either by using part-of-caption as in \modelname or text tokens, as in OVR and LocOV, achieve better attribute mAP than those methods that use a single representation of the text for matching the image.
Interestingly, methods that perform well on object detection are not necessarily better on the overall \taskname.

\paragraph{\modelname ablation}
Table \ref{tab:text_ablation} breaks down the contributions of the parts-of-captions as proxy-labels to the performance of \modelnamenospace{}.
We find that using parts-of-caption as labels helps the model segregate the caption information, improving both the object and attribute detection performance. 
Training the model using noun complements makes the attribute supervision more explicit and makes the best use of the compositionality of the language structure. 
\subsection{Foundation Models Applied to Attributes}
To demonstrate the value of an attribute evaluation benchmark, we tested the zero-shot performance of five pre-trained vision-language models on attributes. 
To focus on attributes, we use the box-oracle setting. We crop the objects using their ground-truth bounding boxes and evaluate the attribute detection for each object instance independently.
Our selection of models was based on the availability of code and model weights.
Moreover, we selected models that process the text and the image independently, such that the matching score can be computed using the cosine similarity between the two representations.

All methods in Table~\ref{tab:ova_sota_box} contain two transformer models that process image and text independently and use the image-text contrastive learning (ITC) loss to learn from image-text pairs.
ALBEF~\cite{albef}, BLIP~\cite{blip}, and X-VLM~\cite{xvlm} additionally include a cross-attention model and use the image-text matching (ITM) loss.
ALBEF and X-VLM use the masked language modeling (MLM) objective~\cite{bert} to predict masked tokens from the caption in a bidirectional manner.
BLIP uses the language modeling (LM) objective~\cite{mnih2008scalable} to generate the caption conditioned on the image in an autoregressive manner.
All three methods use a combination of clean and noisy data for training. ALBEF learns from noisy data by generating pseudo-targets via an online ensemble model~\cite{tarvainen2017mean}.
BLIP instead filters noisy data and generates new captions to learn the multimodal matching. X-VLM uses localized region-text pairs to learn the vision-language alignment at multiple granularities.


\paragraph{Results and discussion}
Table \ref{tab:ova_sota_box} shows the results of foundation models on zero-shot attribute detection. Three interesting behaviors become evident. 

a) Attribute detection is a challenge for foundation models. Compared to zero-shot image classification, where foundation models report very good accuracy~\cite{blip, albef, xvlm, clip, open_clip}, the absolute performance on attributes is surprisingly low. For reference, we trained a supervised attribute detector via cross-validation on our evaluation dataset, which achieved 48.16{\small$\pm$0.52} mAP despite using a small training dataset; see supplementary. Based on the results, foundation models seem to be biased toward object classes and do not pick up fine-grained aspects such as attributes.

b) Not only the quantity but also the quality of training data is important. When scaling from 400M to 2B image-text pairs, OpenCLIP improves by 6.25\% for \textit{All} attribute performance. BLIP improves by 7.06\% when scaling it from 14M to 129M, and quadrupling the data improves X-VLM by 8.46\%. However, the models only reach a good performance once they are further trained on curated data using only ITC and ITM objectives. For instance, ALBEF and BLIP improve their \textit{All} attribute performance this way by 37.25\% and 33.52\%, respectively. 

c)  Localized image region-text matching helps vision-language alignment. X-VLM and \modelnamenospace{}-Box use a localized image region-text matching objective compared to the other methods. X-VLM clearly outperforms all other methods, but it reduces its performance by 6.76\%  when fine-tuning for image-caption retrieval. \modelnamenospace{}-Box outperforms foundation models trained on more than 3000 times larger noisy datasets (CLIP and OpenCLIP), and more than 1000 times larger datasets which include the same clean subset (Table~\ref{tab:training_datasets}(1a)) (ALBEF and BLIP in their pretrained version). In Table~\ref{tab:text_ablation} \modelname shows an increase in performance when using parts-of-caption for explicit visual-text matching during training. We believe that the success of both methods comes from the localized alignment between visual and text context, which is partially lost when specializing for image-caption retrieval. 

 \label{sec:discussion}

\section{Conclusion}\label{sec:conclusion}

We studied the ability of vision-language models to recognize attributes. To this end, we proposed the novel open-vocabulary attribute detection (\taskname) task and introduced the \datasetname\ benchmark, a clean and densely annotated object-level attribute dataset for evaluating \taskname\ task. We provided a baseline method that exploits fine-grained information contained in captions, which outperforms OVD models for the \taskname\ task. Finally, we tested the performance of publicly available foundation models on attribute recognition. We found that the performance of these models on attributes stays clearly behind their performance on objects revealing a direction for further research. 

\section*{Acknowledgement}
This work was supported by Deutscher Akademischer Austauschdienst - German Academic Exchange Service (DAAD) Research Grants - Doctoral Programmes in Germany, 2019/20; grant number: 57440921. 
The Deep Learning Cluster used in this work is partially funded by the German Research Foundation (DFG) - 417962828.
\\
We thank our colleagues Philipp Schröppel, Silvio Galesso and Jan Bechtold for proofreading the paper and providing critical feedback. 
We thank all the annotators especially Mariana Sarmiento and Jorge Bravo for their help, time and effort during the annotation of the OVAD dataset.


\clearpage  
\newpage

\appendix

\tableofcontents

\section{\datasetname\ Benchmark}
\subsection{Attribute taxonomy}
\begin{figure*}[t]
\includegraphics[width=0.8\linewidth]{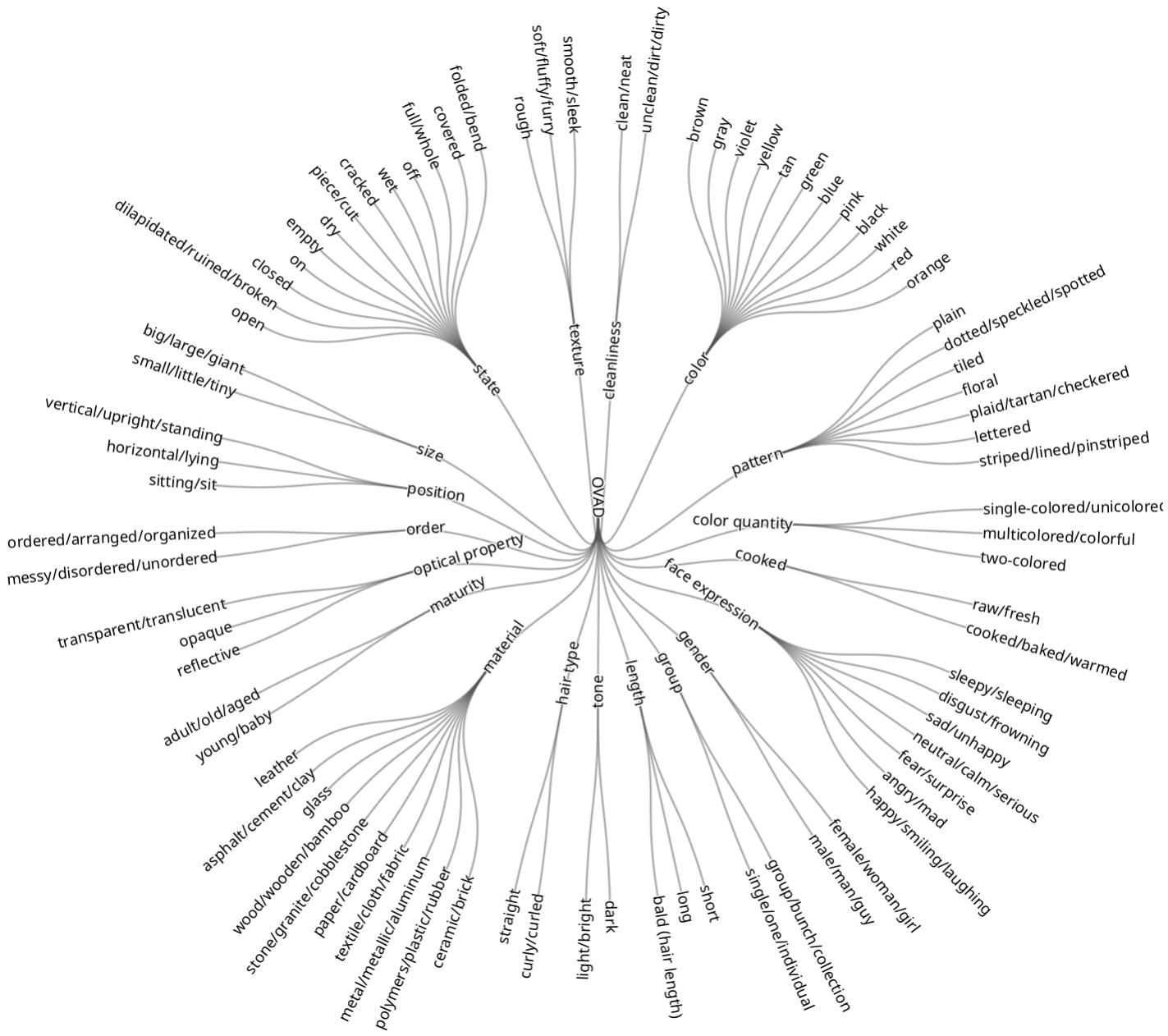}
\centering
\caption{The figure shows the taxonomy of attribute categories as a radial tree. The 117 attribute categories are divided into 19 attribute types, shown in the first circle. Certain attribute types are repeated for the human category, where the color includes hair color and clothes color. Similarly, pattern refers to clothes pattern, length refers to hair length, and tone refers to hair tone. }
\label{fig:att_taxonomy}
\end{figure*}
Figure~\ref{fig:att_taxonomy} shows the attribute taxonomy. We grouped attributes by type to simplify and optimize the annotation process. This diagram corresponds to the attributes of all objects. 19 attribute types are displayed in the inner circle of the radial tree. Each attribute contains its synonyms separated by `/'. For the human category, \textit{color} type refers to \textit{clothes color} and \textit{hair color}; pattern type refers to \textit{clothes pattern}, and length refers to \textit{hair length}. Additionally, we included \textit{sitting} as an attribute to the position type and \textit{bald} as an attribute to the hair length type. In total, we obtain 117 distinct attributes for the \datasetname\ benchmark.
\begin{figure*}[t]
\includegraphics[width=\linewidth]{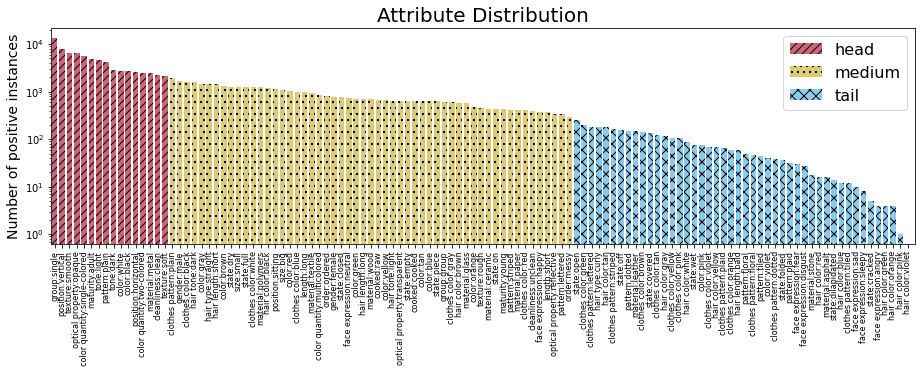}
\centering
\caption{The figure shows the attribute frequency distribution in the \datasetname\ benchmark. Bar colors correspond to the frequency-defined subsets \textit{head}, \textit{medium} and \textit{tail}. }
\label{fig:att_dist}
\end{figure*}

\subsection{Attribute distribution}
Figure \ref{fig:att_dist} shows the long-tailed attribute distribution of positive annotations in the \datasetname\ dataset. Following previous works~\cite{lvis, vaw}, we split attributes in three subsets `head', `medium', and `tail' according to the number of positive instances annotated in the \datasetname\ dataset. To split the classes into `head', `medium', and `tail', we defined two thresholds 
\begin{align*}
    t_{high}& = median(\textbf{f})+std(\textbf{f}), \textrm{and;}\\
    t_{low}& = median(\textbf{f})-std(\textbf{f})/10.
\end{align*}
where $\textbf{f}$ is the frequency vector of the number of positive annotations. `head' corresponds to 15 attribute classes whose frequency is above $t_{high}$, `tail' are the ones below $t_{low}$ composed of 49 attribute classes, and `medium' corresponds to 53 attribute classes, the ones whose frequency is between $t_{high}$ and $t_{low}$.

\subsection{Dataset size analysis}
\begin{figure}[ht]
\includegraphics[width=\linewidth]{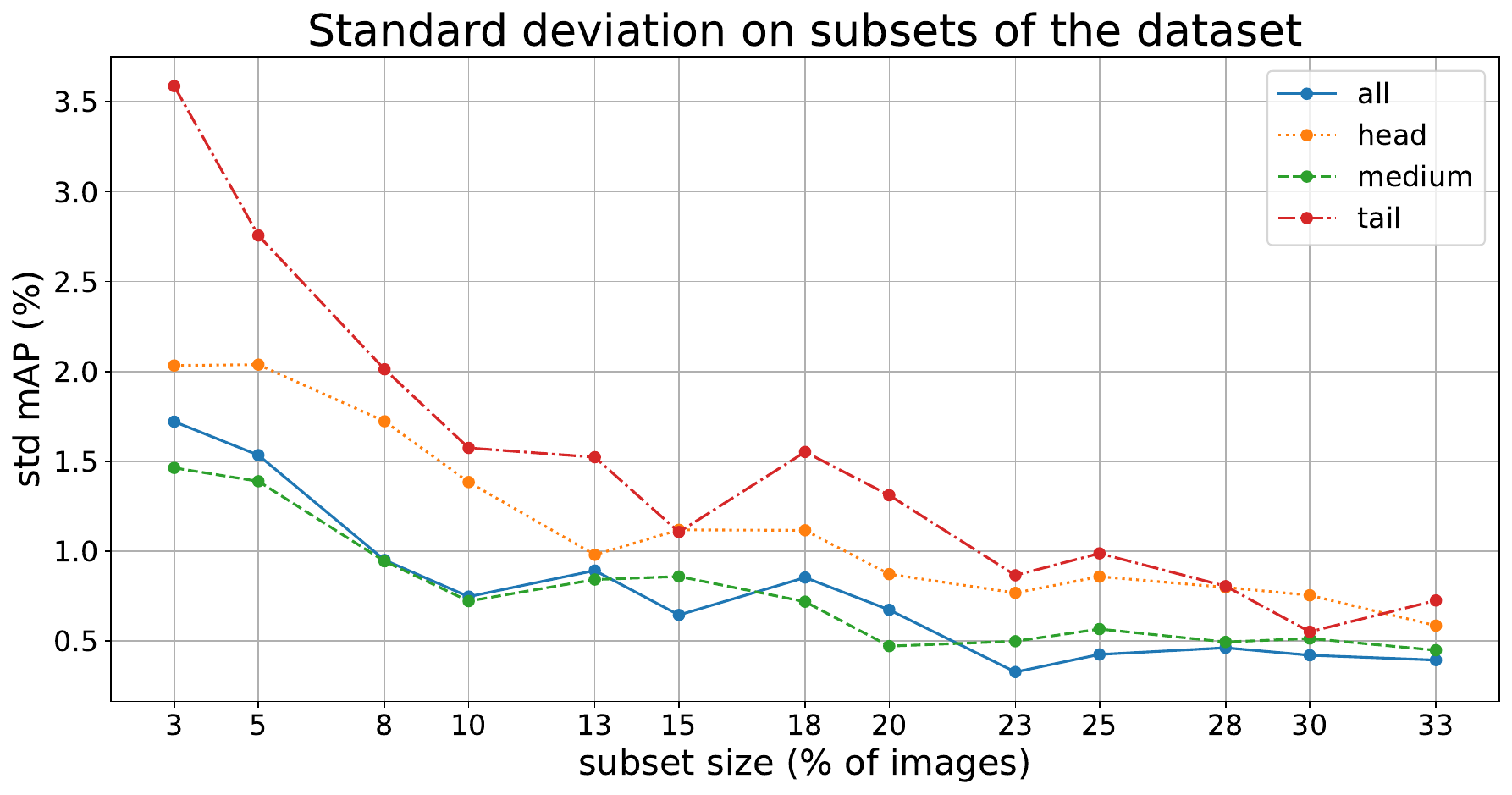}
\centering
\caption{Standard deviation of the mAP performance for the oracle OVAD baseline. We show the scores on differently-sized non-overlapping subsets of images, from 3\% to 33\% of the OVA dataset. All splits (`all', `head', `medium', and `tail') show a decreasing behavior as the number of images increases. At 33\%, the standard deviation is lower than 1\% for all attribute splits.}
\label{fig:dataset_size}
\end{figure}

To show that the size of the \datasetname\ dataset is sufficient for a reliable evaluation of the \taskname\ task, we analyze the standard deviation of the performance of the \modelname-Box. Figure~\ref{fig:dataset_size} shows the standard deviation of the mAP for the frequency-defined subsets: `all', `head', `medium', and `tail'. 
We randomly selected differently-sized subsets of images from the \datasetname\ dataset for this analysis. The size of the subsets range from 3\% to 33\% of the total number of images in the \datasetname\ dataset. We evaluated the \modelname{} model, where ground-truth bounding boxes are provided during evaluation.
The maximum size of a subset is set to 33\% to obtain at least three non-overlapping sets of images, which is required for a reliable calculation of the standard deviation. 
We conducted this experiment six times using different data shuffles to select the splits. In every run we selected a maximum of six non-overlapping splits (for every data size percentage) and computed the standard deviation of the mAP per size of the subset. Then, we average the standard deviation across the six experiments and report the results in Figure~\ref{fig:dataset_size}. 
We observed that the standard deviation decreases as the size of the subset increases. At 23\%, the standard deviation is lower than 1\% for all attribute partitions, and the standard deviation of the `tail' attribute classes is similar to `head' and `medium' attribute partitions. 
When the `tail' attribute curve is extrapolated to 100\% of the dataset size (2000 images), the standard deviation is estimated to be less than 0.3\%.

\section{Dataset Creation}
\subsection{Annotation process}
As mentioned in the main paper, the OVAD dataset is fully annotated by humans following strict guidelines to achieve consistent and dense annotations (guidelines are attached at the end of the supplementary). The annotation process started from scratch for attributes to avoid any pre-existing errors from previous datasets. We utilized the identified 19 attribute types and the taxonomy to facilitate the annotation process. We randomly selected 2000 images from the validation set of MS COCO dataset and used the object annotations as a starting point. 

The annotation system offers a drop-down list of attributes for feasible attribute types for each object instance. For every object, the annotators marked one of the attributes within every attribute type as positive or unknown. For every attribute type, our system allows only one possible attribute selection. We consider all attributes under the same attribute type mutually exclusive except for two types  - \textit{color} and \textit{state}. We use this exclusiveness property to automatically annotate negative attributes by considering all the non-selected attributes (from that attribute type) as negative or unknown. For the attribute type \textit{color}, which is not exclusive, the annotation system offered the possibility to select more than one option as positive. The non-selected colors were either considered negative or ignored depending on the \textit{number of colors} attribute. The type \textit{state} considers a wide range of attributes that are not all mutually exclusive, but some are antonym pairs (\eg wet/dry, open/close). In this case, only the antonyms of the positive-selected attribute were marked as negative, and the rest as unknown. 

It is worth noting that the taxonomy and exclusiveness property was exploited for the benchmark annotation only. It is neither available for training the models nor for predicting the attribute scores, which is done in an open-vocabulary fashion. Providing the attribute classes or the taxonomy to the model during training is against the purpose of the proposed benchmark.

\subsection{Annotation quality control}
We followed a progressive annotation approach. Each annotator received an initial set of images along with the annotation guidelines. Then, a second annotator revised the same set and, based on the annotation guidelines, corrected and completed the missing annotations. Once the annotations were revised, the first annotator received feedback. We repeated this process until a reasonable quality of annotations was achieved (approx. five sets of 50 images each). The progressive process resulted in high annotation quality, with a revision of approximately 80\% of the annotations. The remaining 20\% of the annotation had only one annotation round, corresponding to the last sets annotated by the trained annotators. 

To test the annotation quality of the above-mentioned revised and remaining set of images, we selected 10\% of the data from each of the two sets to perform a second independent annotation from scratch by the experienced annotators and measured the consistency of annotations. As a result, we obtained an overall consistency of 89.44\% for the revised images and 86.35\% for the remaining non-revised set. Additionally, we considered a golden set of 50 images which all the annotators had to do at the end of the annotation process. For this set we obtained an overall consistency of 91.26\%$\pm$2.79. This consistency metric includes positive, negative, and unknown annotations.

\subsection{Human bias in attribute annotation} 
Attribute-level annotations are prone to human biases, which can cause ambiguous type errors, especially in the case of unclear images. We make extensive quality checks to minimize such errors. At least 80\% of the images were revised by a second annotator to establish consistency and correctness of the annotations. On average, annotators spent 12 minutes per image annotating all attributes that apply. While revising, annotators spent approximately 3 minutes per image. The average hourly wage was 12 Euros per hour.  
Our dataset was annotated by fourteen annotators from 6 nationalities, different age groups, sex, and skill levels. This ensures our annotations are balanced for cultural, age, and sex biases.

\subsection{Exceptions in attribute annotation}
Our annotation process offers restricted options to annotate based on the object class category. It does not allow infeasible annotations for each attribute category. \egupper, there is no attribute annotation for the `material' of a person or `cooked' state for a skateboard since it is irrelevant in most of the cases. However, there can be exceptions such as a photo of a person on a banner or a cake in the form of a skateboard. Some of such exceptions are also missed due to stereotyping. \egupper, if there is a car, the annotator might label its material as metal even when it is visually indiscernible. One of the limitations of this work is that our annotation process does not consider such exceptions, thus adding some noise to our annotations. 

\begin{figure*}[h]
\centering
\begin{tabular}{l@{\hspace{1mm}}l}
\includegraphics[trim={0 0cm 0 0cm}, clip, width=70mm]{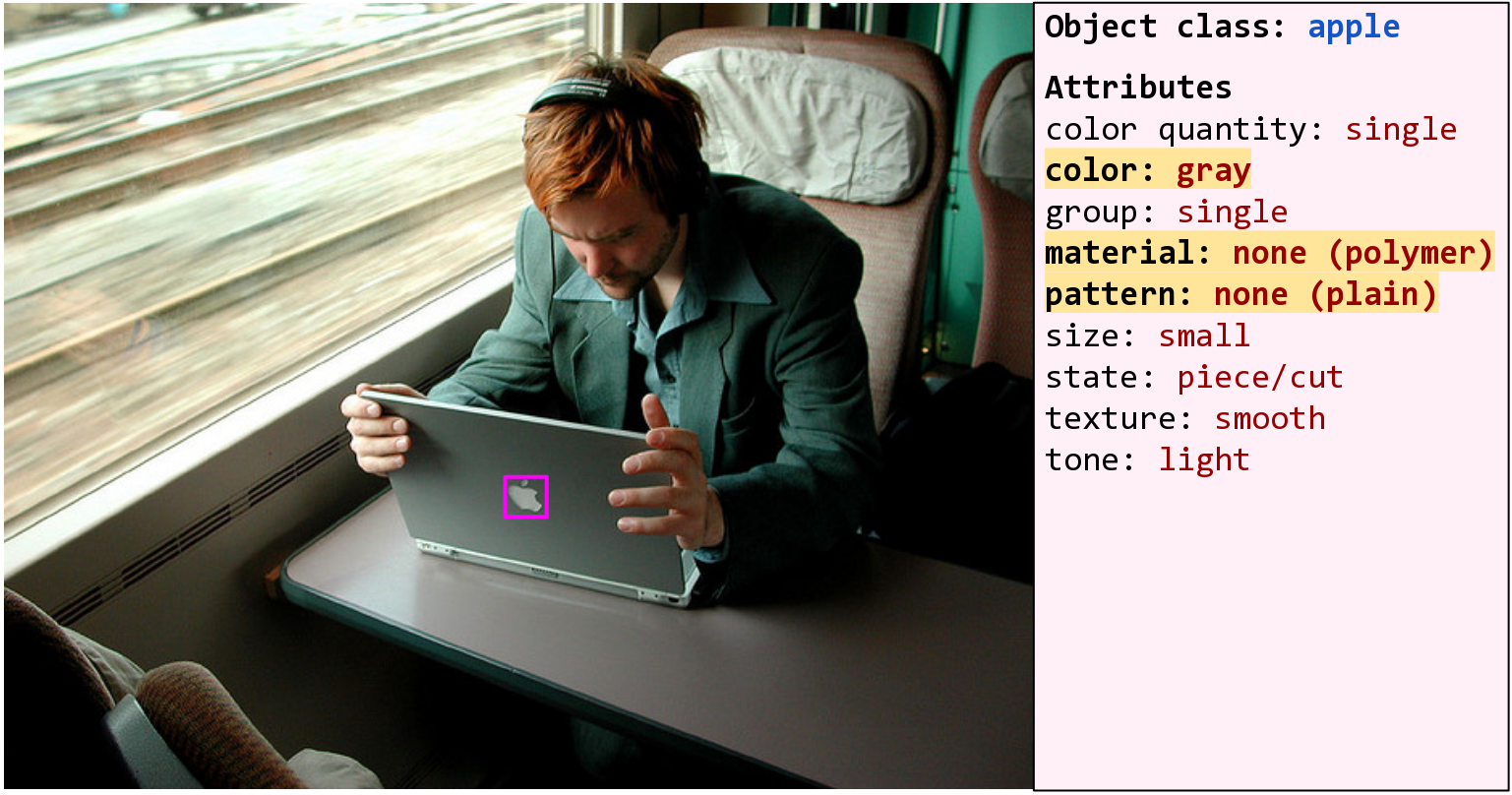} &
\includegraphics[trim={0 0cm 0 0cm}, clip, width=70mm]{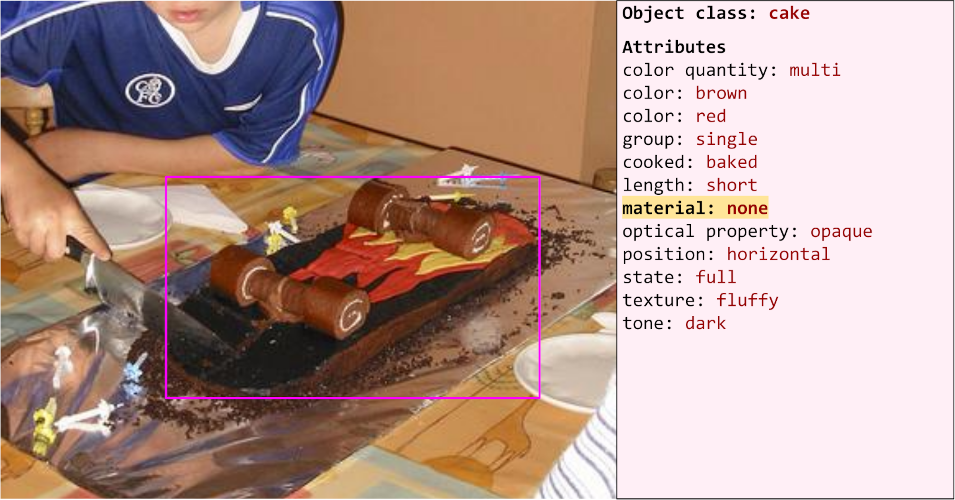}  \\

\includegraphics[trim={0cm 0cm 0cm 0cm}, clip, width=70mm]{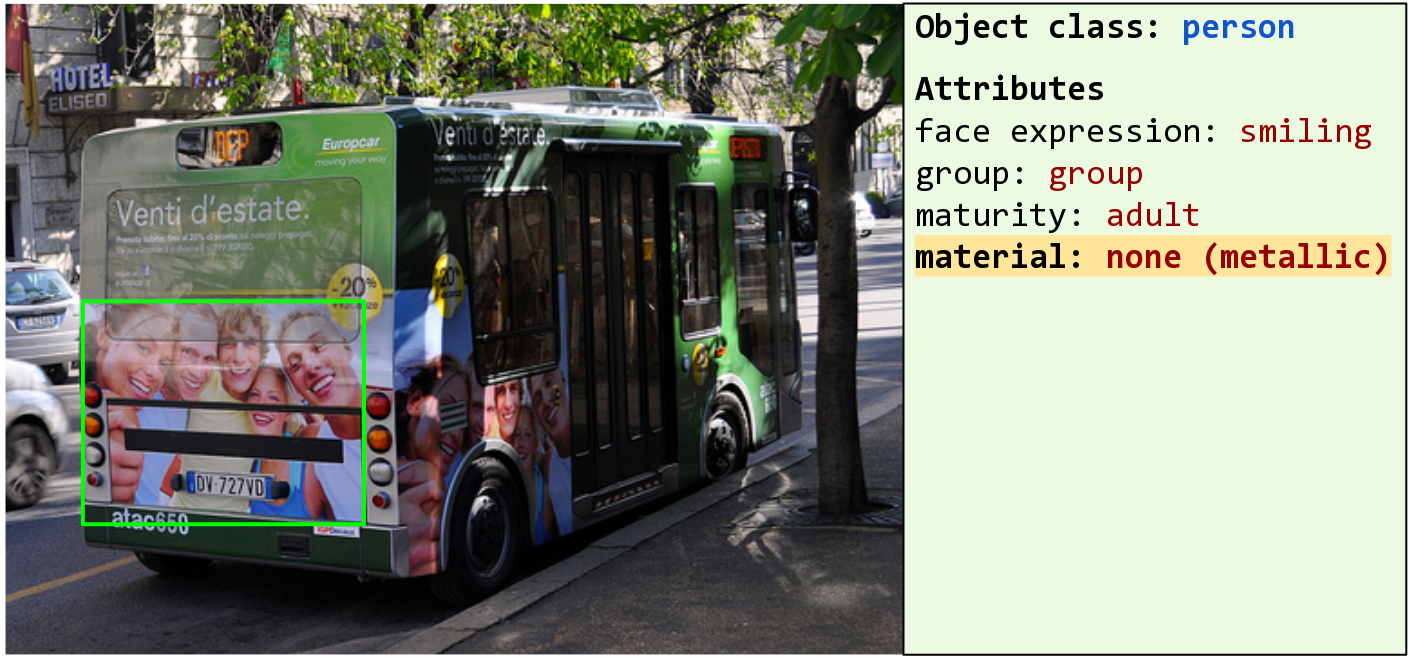}  &
\includegraphics[trim={0 1.2cm 0 0cm}, clip, width=70mm]{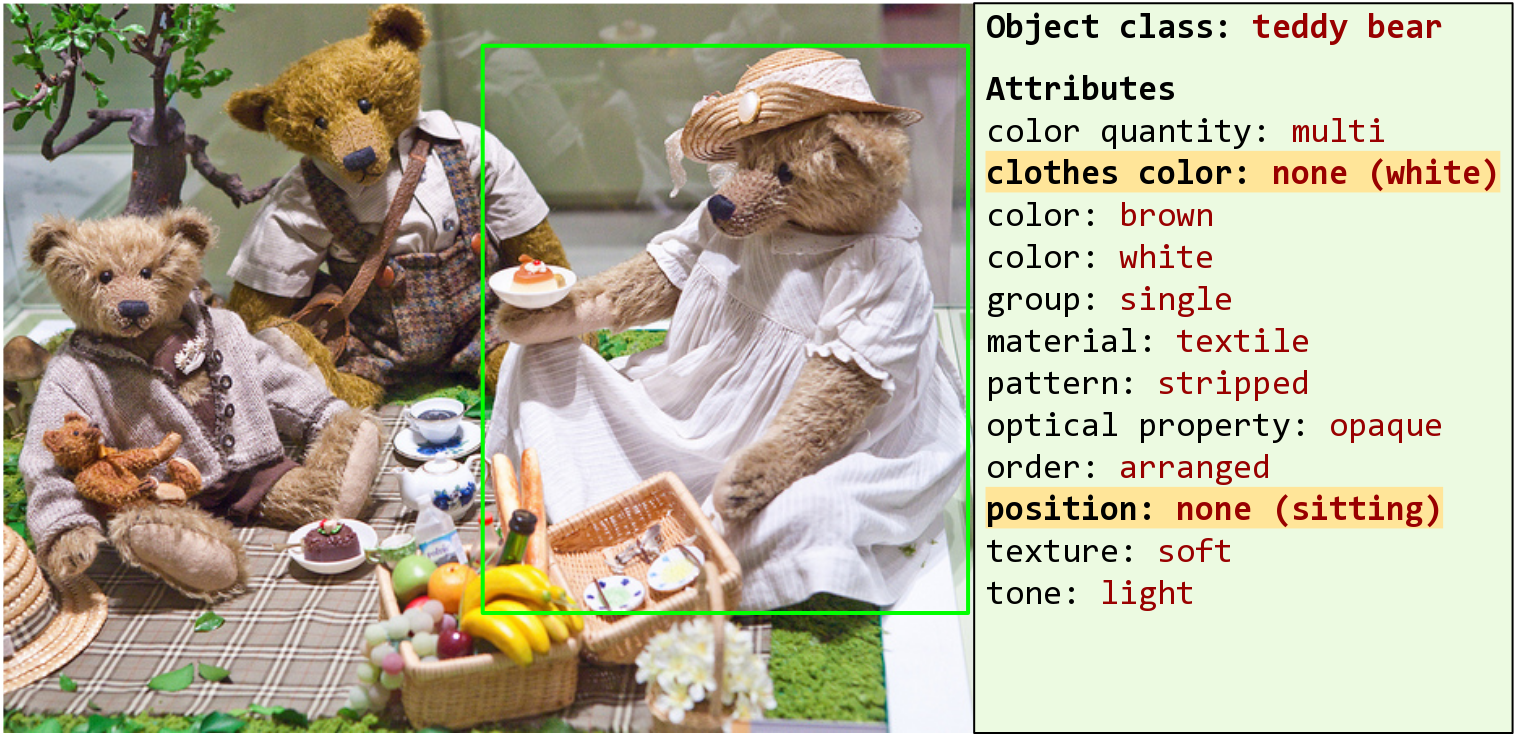}  \\

\includegraphics[trim={0cm 0cm 0 0cm}, clip, width=70mm]{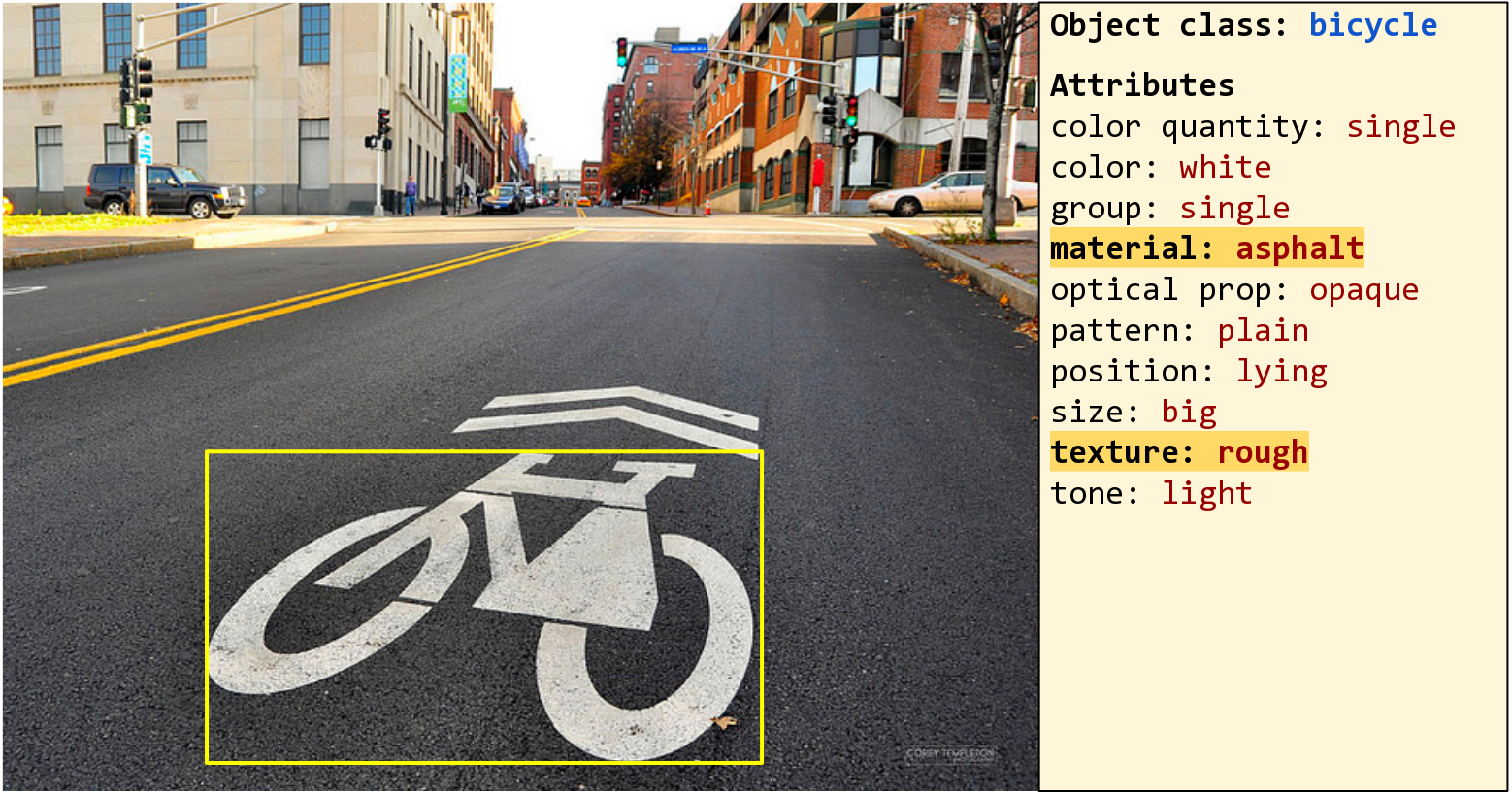} &
\includegraphics[trim={0 0cm 0 0cm}, clip, width=70mm]{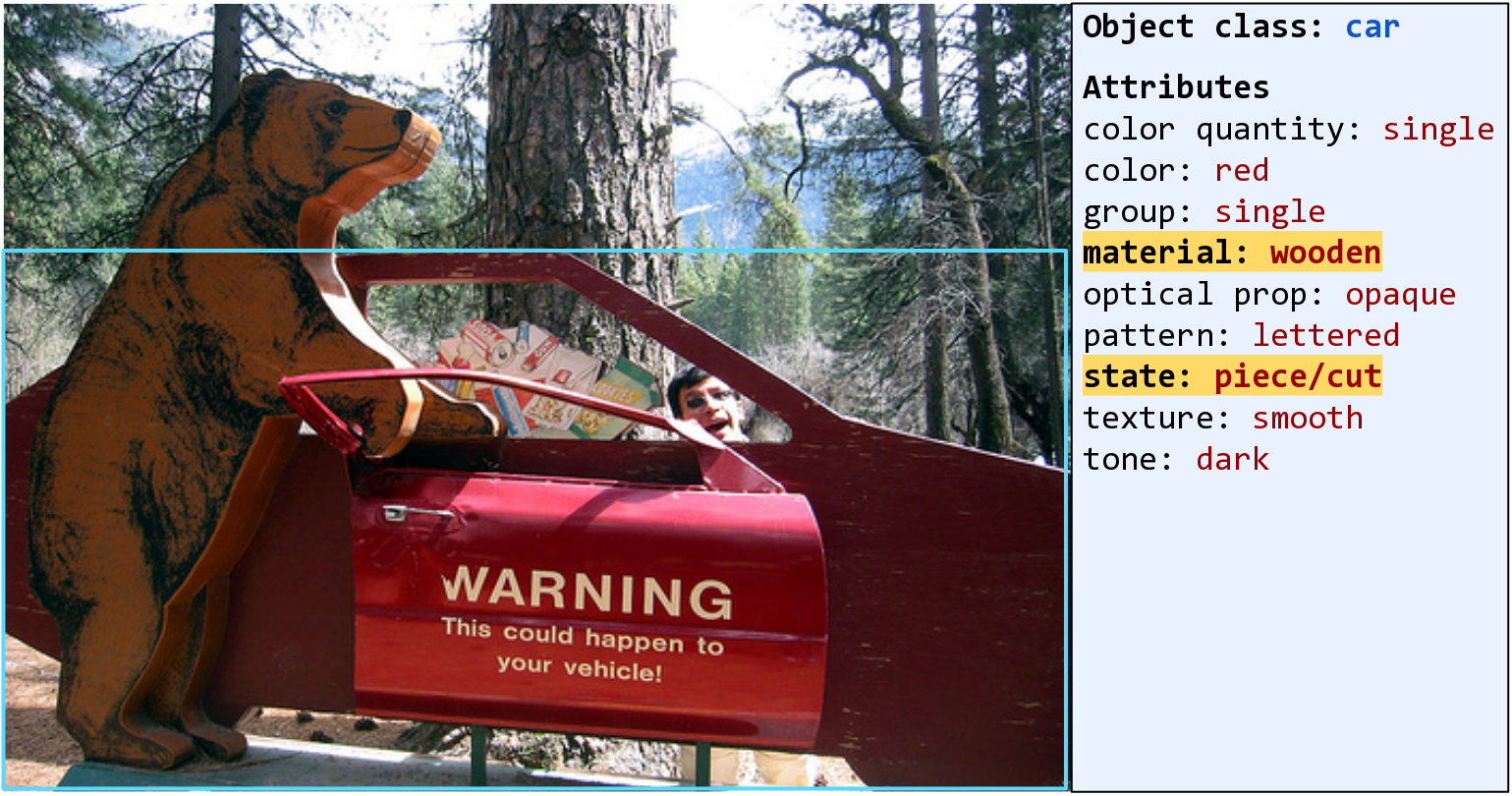}  \\

\includegraphics[trim={0 0cm 0 0cm}, clip, width=70mm]{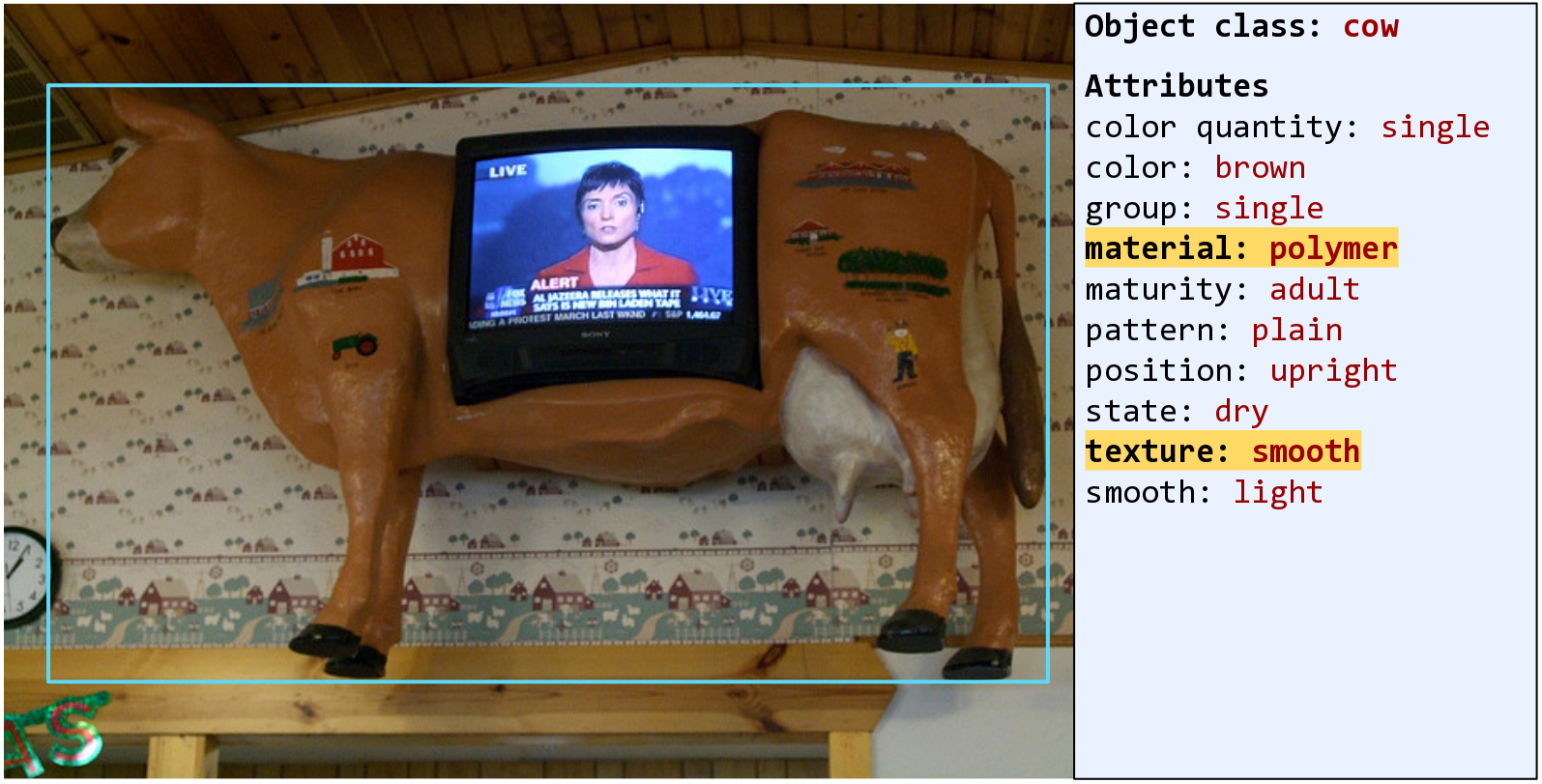} &
\includegraphics[trim={0cm 0cm 0 0cm}, clip, width=70mm]{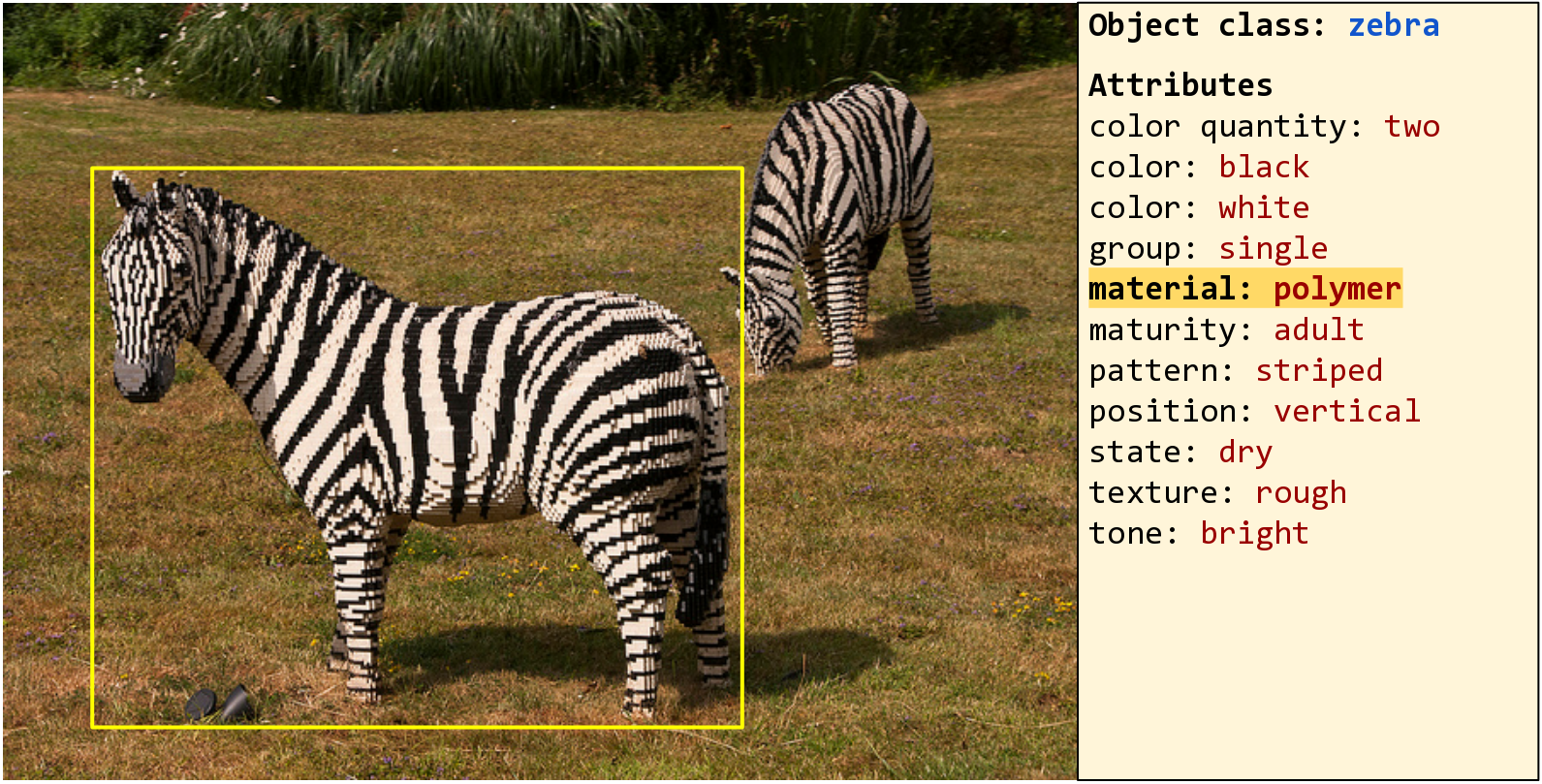}

\end{tabular}
\caption{Exceptions in attribute annotation. Each example shows some of the corner cases present in the OVAD benchmark. The exception attributes for each instance are highlighted in yellow. The correct version is included in the parenthesis if the annotation is marked as unknown in our benchmark.
}
\label{fig:corner}
\end{figure*}

Figure \ref{fig:corner}  shows some examples of exceptions and corner cases in which some attributes are missed due to our annotation system. For every image, the highlighted attributes correspond to the exception cases. For the first and second row, there is a limitation of not considering specific attributes for different objects such as `material' for `apple', `person', or `cake', and `clothes color' for `teddy bear'. For other cases, our annotation system includes the corner cases and selects the correct attribute. The third and fourth row of Figure \ref{fig:corner} shows the possibility of selecting `material' for animals or a different material for some vehicles. 

\subsection{Dataset annotations and visualization}
We included the annotations in a json file in the supplementary material. The format of the annotations is compatible with the MS COCO annotations. Attribute annotations for every instance correspond to a list under the key ``att\_vec" with values of 1, 0, and -1 corresponding to positive, negative and unknown labels respectively. The attribute list is also included in the json file. Additionally, we included a video in the supplementary material showing a demo of our visualization web page. It includes a search system by object and attribute. We distinguish between base and novel object classes and positive, negative and unknown attribute classes using color codes. 

The dataset documentation, annotations, evaluation code and visualization will be publicly available upon acceptance.

\section{Supervised Ablation}
\subsection{OVAD supervised training ablation}

To check the feasibility of the \taskname\ task using our dataset, we perform a supervised 4-fold cross-validation experiment to get an upper bound performance. 
For each run, we consider the 500 image set as the test set and fine-tune a ResNet50~\cite{resnet} architecture pre-trained on Imagenet~\cite{deng2009imagenet} using the remaining 1500 images.
We train the multi-label attribute classification model in the box-oracle setup.
The model achieves an average performance of 48.16{\small$\pm$0.52} mAP compared to the chance performance of 8.29{\small$\pm$0.06} mAP.
For reference, our \modelname-Box achieves 23.30{\small$\pm$0.76} mAP on the same splits. 
\subsection{Cross-dataset transfer ablation}
Our primary interest lies in the OVAD task, which considers all attributes as novel categories. However, in order to investigate the potential transfer of knowledge from previous benchmarks to our OVAD benchmark, we conducted an ablation experiment. We trained two ResNet50 networks using the cropped objects from the COCO Attributes~\cite{coco_attributes} and VAW~\cite{vaw} datasets, respectively. We trained a multi-label attribute classification model using the box-oracle setup, incorporating a projection layer at the end of the network to obtain vector representations of the same dimension as the CLIP text encoder. We computed the similarity between every attribute encoded by the CLIP text encoder and the object visual vector. We train the models using binary cross entropy loss with the positive and negative attribute labels from the datasets. Our models achieved a performance of 15.96 mAP and 18.20 mAP on OVAD (box-oracle setting) after training on the COCO Attributes and VAW datasets respectively.

\section{\modelname}
\subsection{Implementation details}
Our \modelname\ method uses ResNet50~\cite{resnet} as backbone, pre-trained on ImageNet \cite{deng2009imagenet}, for the detector model $F$ and the CLIP text encoder~\cite{clip} as the language model $G$. Similar to CLIP, we use the cosine similarity (Equation (1) in the main paper) between the visual representation $f_b$ of the object's bounding box and the text class embedding $g_c$ for applying the classification losses during training and for calculating the prediction scores during inference.

We set $\tau$ to 50 during training and testing for both object and attribute prediction. The temperature parameter is selected empirically for object detection. For reference, Detic~\cite{detic} uses a value of 50, and CLIP~\cite{clip} uses a value of 14.29 for the temperature hyperparameter. As mentioned in Section 4 in the main paper, we use a binary cross entropy objective for all the losses that use the classification head, which are image-caption matching and parts-of-caption matching (noun/noun phrase/noun complement) with max-area box. 
For efficiency, we compute the text representations offline and load the features of positive and negative image-text pairs during training. We select one positive and 63 negative captions for image-caption matching and compute the similarity with the bounding box covering the whole image. For parts-of-caption matching, we select all positive samples $p$ and $50-p$ negative parts-of-caption to compute binary cross-entropy with the maximum area bounding box proposal.

During training, we use a base learning rate of 0.02 with a step reduction of 10x at 60\,k and 80\,k iterations. We train the model for a total of 90\,k iterations with 1\,k warmup steps using the SGD optimizer, similar to previous open-vocabulary detection methods~\cite{ovr_baseline, detic}. The training is done per batch of one type of data at a time; a batch contains either box+class labels of base classes or caption+parts-of-caption labels. We use the same sampling ratio of training as Detic~\cite{detic} of 1:4 for the batch type of images, using four times more batches with the captions. We use a batch size of 64 for image-caption data and 16 for box+class data. 

During inference, we considered several synonyms for every attribute class representing the category. We refer to this set as $S_a = \{w_i: w_i$ is a synonymous term for attribute $a\}$. These sets are shown on Figure~\ref{fig:att_taxonomy} and are listed under every attribute type using `$/$'. When calculating the text-attribute embedding $g_a$ for every category, we average the representations of the individual synonyms using
\begin{equation}
    g_a = \frac{1}{|S_a|}\sum_{w_i \in S_a} g_{w_i}.
\end{equation}%
We use the similarity score (Equation (1) in the main paper) as the prediction score for every attribute. 
\section{Experimental Extension}

\subsection{Open-vocabulary attribute detection results} 
\begin{table*}[t]
\begin{center}
\begin{tabular}{ |c|ccc|ccc|} 
    \hline 
    \multirow{2}{*}{Method} & \multicolumn{3}{|c|}{ Generalized (OVD-80) - 2,000 images } & \multicolumn{3}{|c|}{ Generalized (OVD) - 4,836 images} \\
    & Novel (32) & Base (48) & All (80)
    & Novel (17) & Base (48) & All (65)\\ 
    \hline
    OV-Faster-RCNN & 0.4 & 53.1 & 32.0 & 0.3 & 53.0 & 39.2 \\ %
    OVR~\cite{ovr_baseline}          & 17.9 & 51.8 & 38.2 & 22.8 & 46.0 & 39.9\\
    VL-PLM~\cite{vl-plm}*  & 19.7 & \textbf{58.9} & 43.2 & 34.4 & \textbf{60.2} & \textbf{53.5}\\  
    Detic~\cite{detic}*       & 20.0 & 49.2 & 37.5 & 27.8 & 47.1 & 45.0\\ %
    LocOv~\cite{locov}        & 22.5 & 52.5 & 40.5 & 28.6 & 51.3 & 45.7\\
    \modelname   & 24.7\std{0.6} & 49.1\std{0.2} & 39.3\std{0.4} & 30.0\std{0.5} & 48.3\std{0.4} & 43.5\std{0.3} \\ %
    Rasheed \etal~\cite{bridging}*  & \textbf{32.5} & 56.6 & \textbf{46.9} & \textbf{36.6} & 54.0 & 49.4\\ %
    \hline
\end{tabular}
\caption{AP\textsubscript{50} on Open-Vocabulary Object Detection. *: novel class labels were used during training to filter captions and obtain the image tags (Detic), or to obtain pseudo-labels (VL-PLM).}
\label{tab:coco_sota_gen}
\end{center}
\end{table*}

Table \ref{tab:coco_sota_gen} shows the results of evaluating the different state-of-the-art methods on both open-vocabulary object detection benchmarks based on the MS-COCO~\cite{coco} validation dataset. We use two different sets of image annotations, our extended OVD-80 benchmark with updated object annotations and the OVD benchmark proposed by Bansal~\etal~\cite{bansal2018zero}. Results are shown in the \textit{Generalized} scenario where detection is performed across both base and novel classes together. 

Our method achieves a high AP\textsubscript{50} performance for novel categories with a trade-off with base class performance. Methods marked with * use novel classes to obtain pseudo image-labels, therefore are not open-vocabulary by definition. The order of the methods is consistent across both image sets, OVD and OVD-80. OVD-80 has 32 novel objects class making the object detection task more challenging compared to having 17 novel object classes in OVD. 

\subsection{Performance per attributes type}

\begin{figure*}[t]
\centering
\includegraphics[ width=0.9\textwidth]{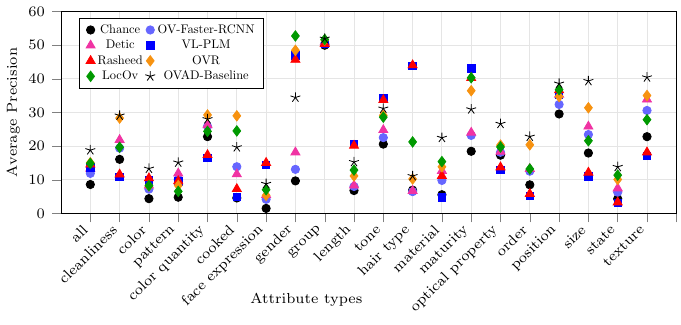}
\caption{Comparison between different baseline methods on open-vocabulary attribute detection on the \datasetname\ benchmark. }
\label{fig:method_comparison}
\end{figure*}

\begin{figure*}[t]
\centering
\includegraphics[ width=0.9\textwidth]{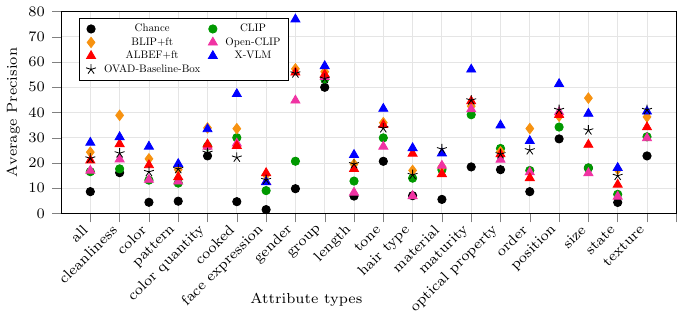}
\caption{Comparison between different the different foundation models on the box-oracle \datasetname\ benchmark. }
\label{fig:vlm_method_comparison}
\end{figure*}
Figure \ref{fig:method_comparison} shows the performance of the five methods for every type of attribute category. Categories such as material, optical property, order, size, and texture show a bigger improvement over chance performance than other attribute types.

Figure~\ref{fig:vlm_method_comparison} shows the mAP scores for all six foundation models per attribute type in the box-oracle setup. X-VLM outperforms all other methods by a large margin for the majority of attribute types. Some attributes such as cooked, gender and maturity have a higher relative improvement over chance level. 

\section{Qualitative Results}

\begin{figure*}[h]
\centering
\begin{tabular}{l@{\hspace{1mm}}l}
 \includegraphics[trim={0 0cm 0 0cm}, clip, width=58mm]{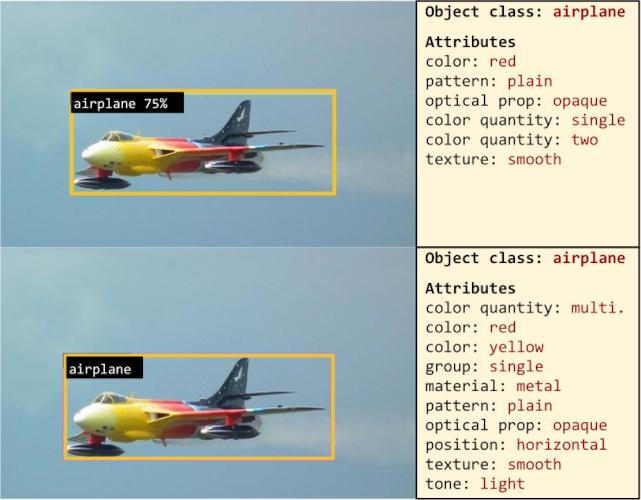} &
 \includegraphics[trim={0 0cm 0 0cm}, clip, width=76mm]{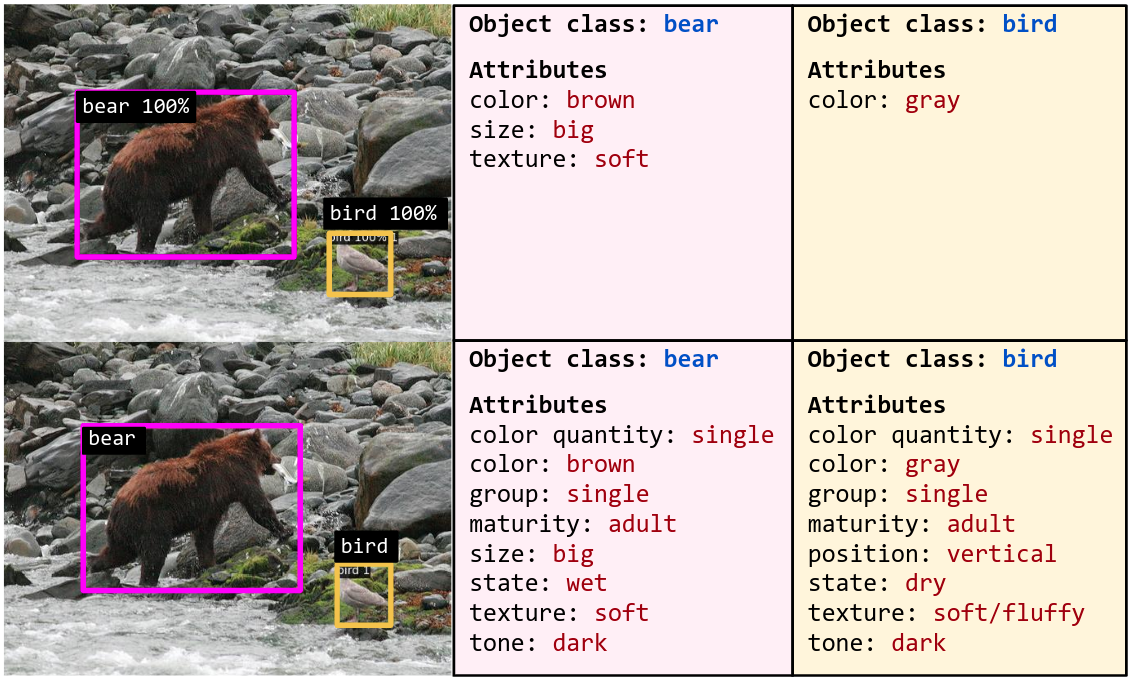}  \\

 \includegraphics[trim={0 0cm 0 0cm}, clip, width=58mm]{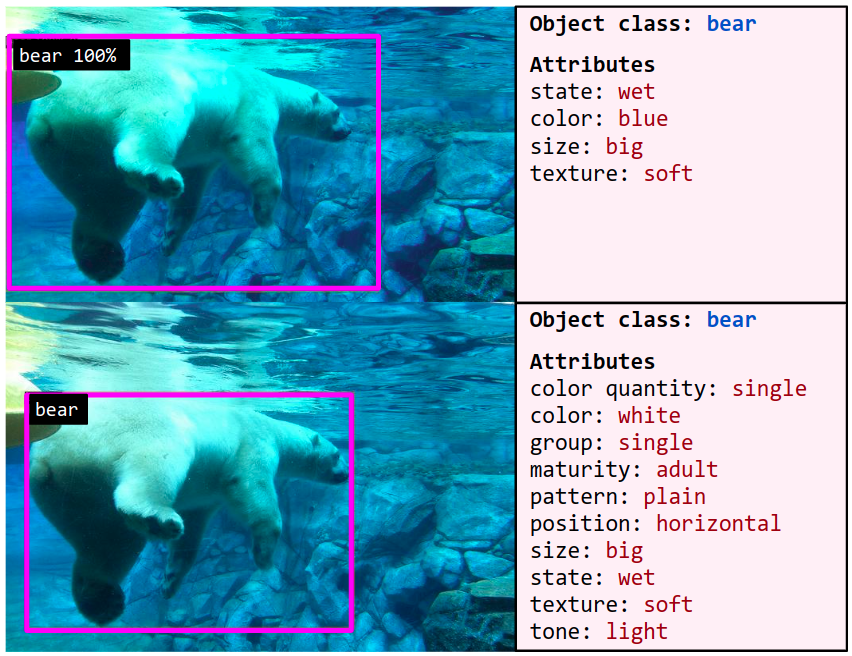} &
 \includegraphics[trim={0 0cm 0 0cm}, clip, width=76mm]{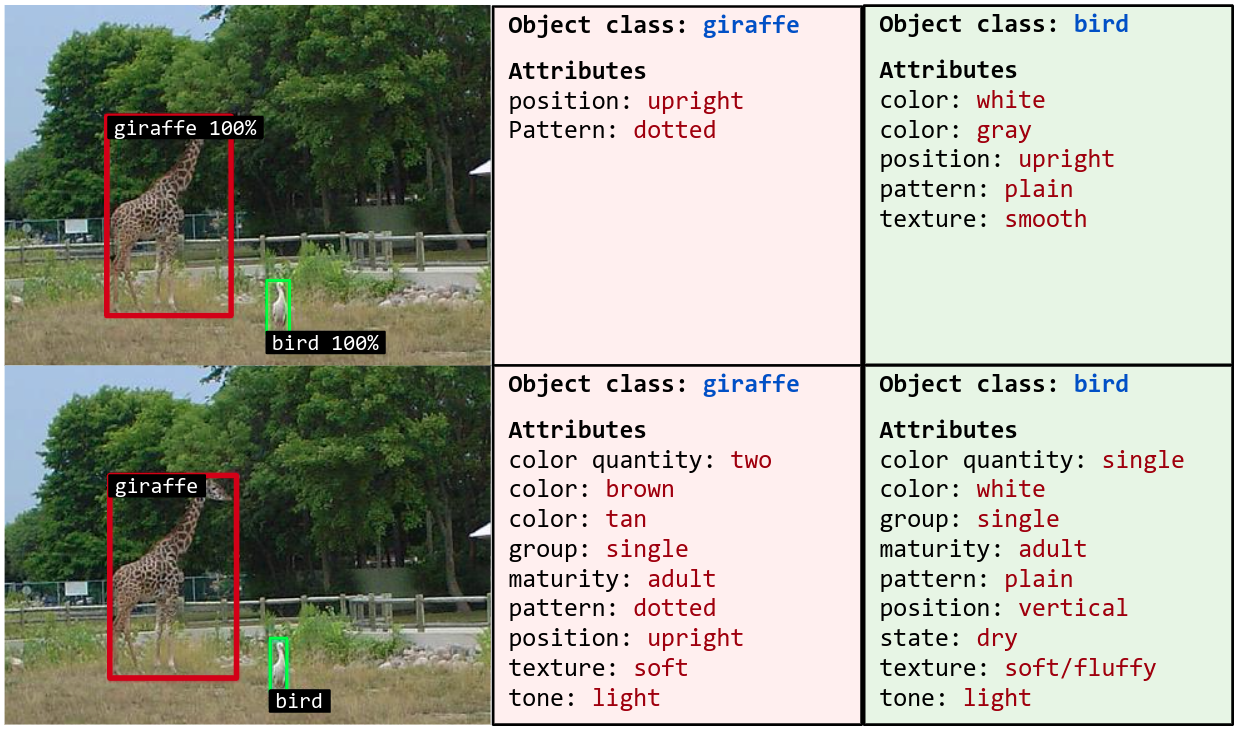} \\
 
 \includegraphics[trim={0 0cm 0 0cm}, clip, width=58mm]{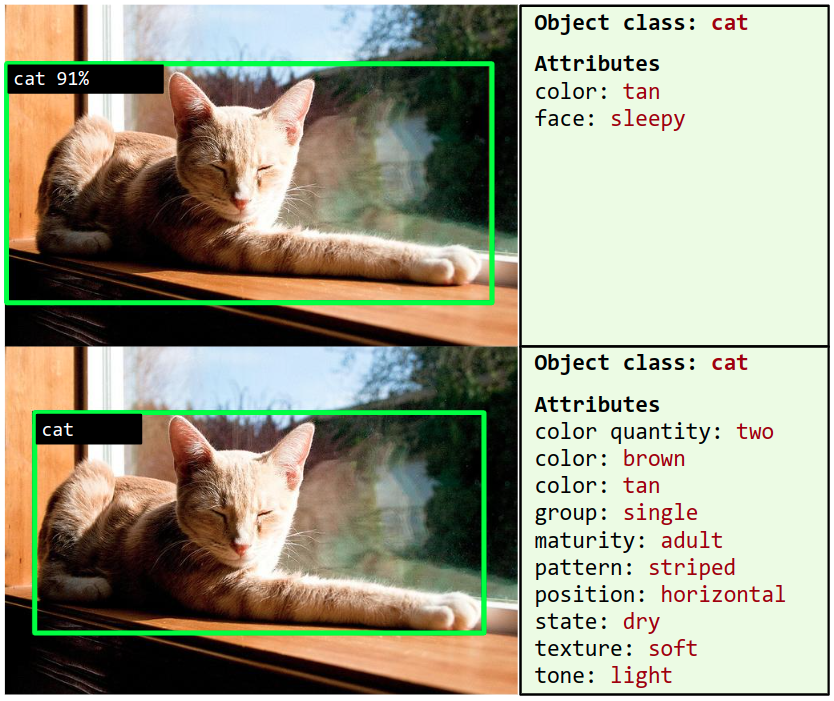} &
 \includegraphics[trim={0 0cm 0 0cm}, clip, width=76mm]{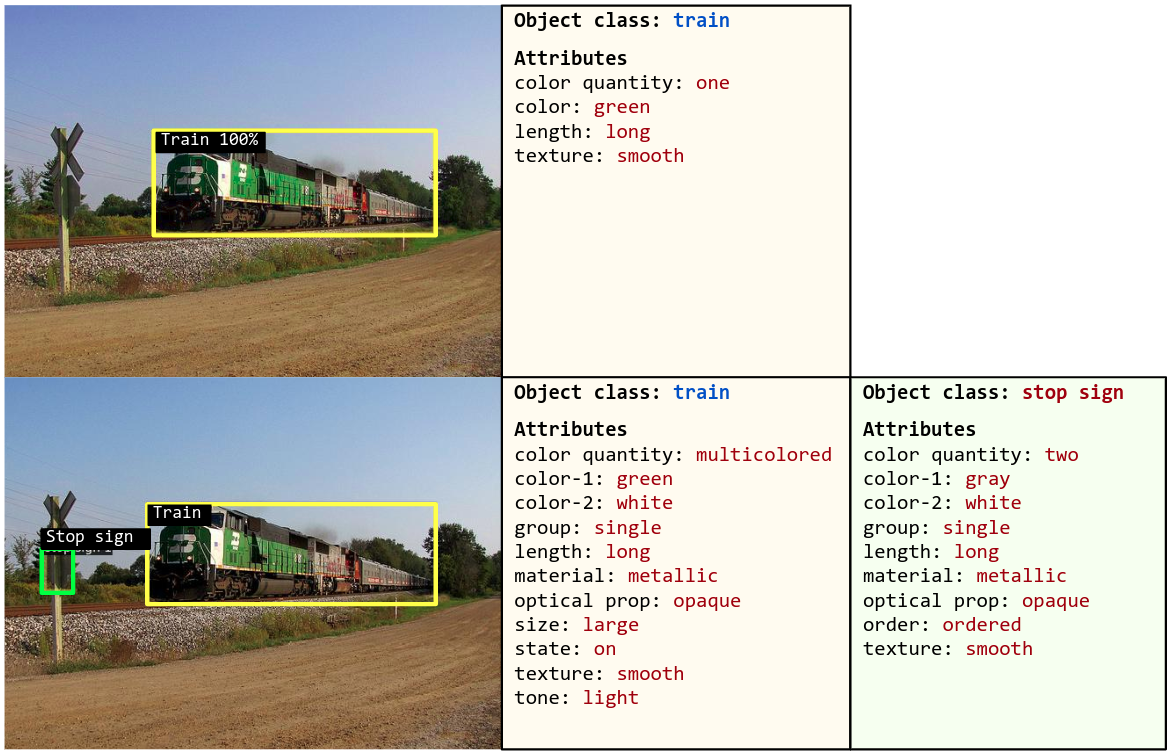}\\
 
 \multicolumn{2}{c}{ \includegraphics[trim={0 0cm 0 0cm}, clip, width=136mm]{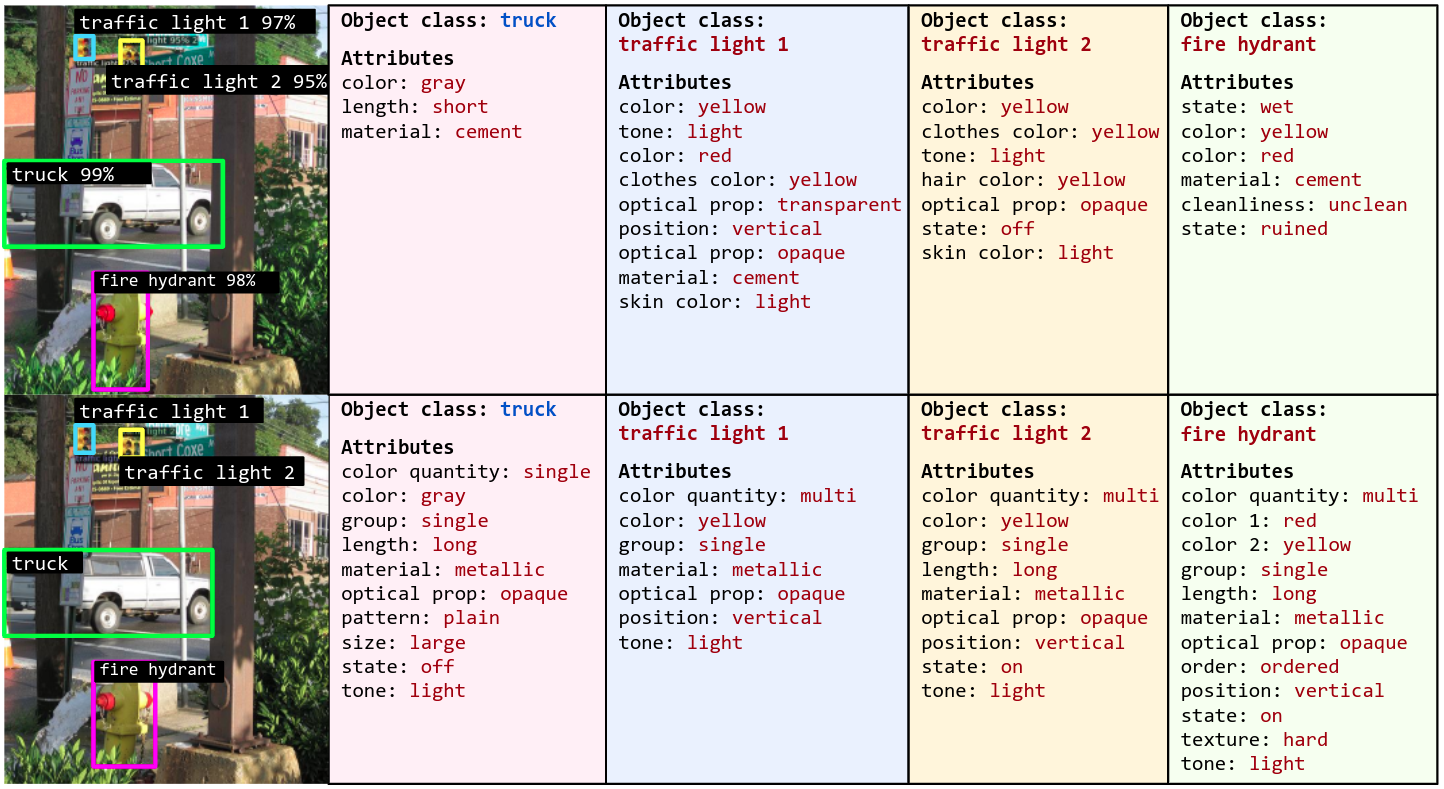}}
 
\end{tabular}
\caption{Qualitative examples. For each example, top row shows the predictions of the proposed OVAD base model and bottom row shows the ground-truth.}
\label{fig:qualitative}
\end{figure*}



 
 
 
Figure \ref{fig:qualitative} shows qualitative examples of the baseline method for OVAD. For every example, the first image corresponds to the prediction and the second to the ground truth annotations. All \textcolor{blue}{base} and \textcolor{dred}{novel} entities are shown in \textcolor{blue}{blue} and \textcolor{dred}{red} respectively. 
The prediction which has the maximum overlap with the ground truth bounding box is considered as the final prediction. 
We rank the attribute prediction scores for each attribute category and select the top 200 scores for visualization. Figure~\ref{fig:qualitative} shows some images with high mAP performance.

\section{Licences of Assets}
We provide some additional details about the datasets, codes and other used assets. These details include the source and their licenses.

\paragraph{(CVAT) Computer Vision Annotation Tool}
The application and the code for the CVAT tool~\cite{cvat} are available at the GitHub repository: \url{https://github.com/openvinotoolkit/cvat}, web page: \url{https://cvat.org}. The repository is licensed under the MIT license.

\paragraph{MS COCO}
Both MS COCO detection and caption datasets \cite{coco, cocoCaptions} are available at their web page \url{https://cocodataset.org} and github repository \url{https://github.com/cocodataset/cocoapi}. These dataset follow the following licences: Attribution-NonCommercial-ShareAlike License, Attribution-NonCommercial License, Attribution-NonCommercial-NoDerivs License, Attribution License, Attribution-ShareAlike License, Attribution-NoDerivs License, No known copyright restrictions, United States Government Work.

\paragraph{OVAD dataset attribute annotations license}
The OVA benchmark, annotations along with the website are licensed under a \href{https://creativecommons.org/licenses/by-nc-sa/4.0/}{Creative Commons Attribution-NonCommercial-ShareAlike 4.0 International License}.


\section{Annotation Guidelines}
These guidelines were provided to the annotators to maintain consistency and agreement in the annotations. 
\begin{enumerate}
    \item Given an image, check for every object marked and verify that it has the correct class.
    \item Add bounding boxes for the missing objects, revise inaccurate bounding boxes. 
    \item Annotate attributes as positive only based on their visual appearance. 
    \item  Assign `unknown' for cases where: (a) the attribute is not visible, like in the presence of occlusion or (b) a discrete label can not be assigned because of ambiguity or an in-between case.
    \item Check through all the feasible attribute types and select the most appropriate attribute category as positive according to the attribute descriptions (included below).
\end{enumerate}


We considered four types of object categories. The valid set of attribute types is allocated based on this object category.   \\ 
\begin{itemize}
    \item human: person
    \item animal: bird,  cat,  dog,  horse,  sheep,  cow,  elephant,  bear,  zebra,  giraffe
    \item food: banana, apple, sandwich, orange, broccoli, carrot, hot dog, pizza, donut, cake
    \item object: bicycle, car, motorcycle, airplane, bus, train, truck, boat, traffic light, fire hydrant, stop sign, parking meter, bench, backpack, umbrella, handbag, tie, suitcase, frisbee, skis, snowboard, sports ball, kite, baseball bat, baseball glove, skateboard, surfboard, tennis racket, bottle, wine glass, cup, fork, knife, spoon, bowl, chair, couch, potted plant, bed, dining table, toilet, tv, laptop, mouse, remote, keyboard, cell phone, microwave, oven, toaster, sink, refrigerator, book, clock, vase, scissors, teddy bear, hair drier, toothbrush
\end{itemize}
Based on these categories, we defined the possible attributes to assign from the 19 attribute types according to attribute descriptions. \\
\paragraph{Attribute categories}
\begin{enumerate}
    \item \textbf{cleanliness} 
    \begin{itemize}
        \item \textbf{clean/neat -} This attribute is marked when an object is clearly clean. This attribute usually applies to objects that appear to be new or especially clean for the picture and for animals that are fully visible and no dirt can be seen. 
    
        \item \textbf{unclean/dirt/dirty/muddy -} This attribute is marked when an object is clearly dirty. This attribute usually applies to objects with something over them, some visible spillage or dust. It usually applies to graffiti on walls not designed for that, animals with mud, dirty dishes, or objects on the street that are poorly maintained.
    \end{itemize}
    
    \item \textbf{color,  clothes color,  hair color: (black, white, gray, tan, brown, green, red, yellow, blue, orange, violet, pink) -} This attribute refers to the visible color of the object. Color/clothes color/hair color applies to different object types differently. For example, hair color and clothes color apply only to humans; however, humans have no color attribute. 
    
    \item{\textbf{color quantity}} 
    \begin{itemize}
        \item \textbf{single-color, two-colored -} This attribute is marked when an object comprises exactly one or two colors. 
    
        \item \textbf{multi-color -} This attribute is marked when more than two colors are present in the object. Even when some text or lines are in a third color, it is marked as multi-colored.
    \end{itemize}
    
    \item \textbf{cooked -} This attribute type is marked only for food object categories. It denotes whether the food is cooked/baked or raw.
    
    \item \textbf{face expression -} This attribute type refers to the person's facial expression. This attribute is only marked when the face of the person is clearly visible.

    \item \textbf{gender -} The attribute type refers to the gender of the person. This attribute is marked based on the combination of body features, face features, clothing, context, etc. We understand that sometimes it can be challenging to mark the gender of a person just based on appearance. Therefore, we take extreme measures, particularly for this attribute, and only mark it when it is very evident. 

    \item \textbf{group -} This attribute type refers to the number of instances in the bounding box. There are two possible categories for this type of attribute - single/individual or group/collection. 

    \item \textbf{hair type -} This attribute type refers to the hair type of the person and is classified either as curly/curled or straight.

    \item \textbf{length: long, short; hair length: long, short, bald} This attribute type is marked when an object is evidently extra-long/short relative to its standard/average size. 
    For example, an international airplane like Airbus-380 is marked as long, whereas a private jet is marked as short. 
    
    \item \textbf{material -} This attribute type refers to the most visible material in appearance. If two dominant materials exist, then the structure's material is marked, and if the object is covered with another material, then the surface's material is marked.  
       
    \item \textbf{maturity -} This attribute type refers to the physical maturity of humans or animals. This attribute is either marked as adult/old or young/baby. 
    
    \item \textbf{optical property -} This attribute type expresses the optical property of the object's material. Most objects are marked as opaque. If the surface of the opaque object is reflective, then the optical property is marked as reflective. Other remaining attribute includes transparent/translucent objects.
    
    \item{\textbf{order}} 
    \begin{itemize}
        \item \textbf{unordered -} This attribute is marked when an object is cluttered and fails to follow any particular order. This attribute usually applies to objects which carry or comprise multiple elements or parts. For example, a working desk with cluttered items or a couch with objects lying on it in an unorganized way.
    
        \item \textbf{ordered -} This attribute is marked when the object is organized and holds an order. This attribute usually applies to objects which carry or comprise multiple elements or parts. 
    \end{itemize}
       
    \item \textbf{pattern -} This attribute type refers to the pattern of the surface of the object. It includes clothes patterns, object surface patterns, etc. If the surface is a mixture of two patterns, then the pattern is marked as unknown.
    
    \item \textbf{position -} This attribute type refers to the orientation of the object. This attribute also includes the sitting attribute. There is no consensus on the object's orientation for certain classes like table, bowl, and microwave. Therefore, they are marked as unknown.
         
    \item \textbf{size: big, small -} This attribute type is marked when an object is evidently extra-big/small relative to its standard/average size. For example, an elephant could be considered big by default, however it is only marked as big only if it is extra large relative to a normal-sized elephant.

    \item \textbf{state -} This attribute type is a non-exclusive attribute that contains multiple attribute sub-types like dry/wet, closed/open, turned on/off, etc. Different sub-types apply to different object categories. Electronic devices can be marked as either turned on or off. Animals can be marked as dry or wet. Container-type objects can be marked as either open or closed. 
    
    \item \textbf{texture} This attribute type refers to the visual appearance of the consistency of the surface of the objects. 
    \begin{itemize}
        \item \textbf{smooth/sleek -} This attribute is assigned for objects having a flat, regular surface or appearance.
        \item \textbf{soft/fluffy/furry -} This attribute corresponds to objects with surfaces covered with fur or hair, as well as objects that could be easily pressed and deformed.
        \item \textbf{rough -} This attribute describes objects with irregular or uneven textures whose appearance shows irregularities on the surface.
    \end{itemize}  
           
    \item \textbf{tone -} This attribute type refers to the tone of the surface of the object. The tone is either marked as light/bright or dark. It could refer to the color tone of the object or hair tone, depending on the object class. For example, the tone attribute is not marked for the person's skin. For the person object class, only hair tone is marked. 
    
    
    

    
\end{enumerate}

\clearpage  
\end{document}